\documentclass[preprint,12pt]{elsarticle}

\usepackage{amssymb}
\usepackage{longtable}
\usepackage{booktabs}
\usepackage{amsmath}
\usepackage{amsthm}
\usepackage{multirow}
\usepackage{array}
\usepackage{float}
\newtheorem{definition}{Definition}

\usepackage{color}
\usepackage{xcolor}
\usepackage{subcaption}
\usepackage{graphicx}

\usepackage{algorithm}
\usepackage[noend]{algpseudocode}
\usepackage{url}

\journal{Elsevier}

\begin{document}

\begin{frontmatter}



\title{Heterogeneous Graph Neural Networks with Post-hoc Explanations for Multi-modal and Explainable Land Use Inference
}

\affiliation[label1]{organization={Imperial College London, Department of Civil and Environmental Engineering},
            addressline={Exhibition Rd, South Kensington},
            city={London},
            postcode={SW7 2AZ},
            country={UK}}
\affiliation[label2]{organization={Imperial College London, Department of Computing},
            addressline={
            South Kensington Campus},
            city={London},
            postcode={SW7 2AZ},
            country={UK}}

\author[label1]{Xuehao Zhai}
\author[label2]{Junqi Jiang}
\author[label2]{Adam Dejl}
\author[label2]{Antonio Rago}
\author[label1]{Fangce Guo}
\author[label2]{Francesca Toni}
\author[label1]{Aruna Sivakumar}

\begin{abstract}
Urban land use inference is a critically important task that aids in city planning and policy-making. Recently, the increased use of sensor and location technologies has facilitated the collection of multi-modal mobility data, offering valuable insights into daily activity patterns. Many studies have adopted advanced data-driven techniques to explore the potential of these multi-modal mobility data in land use inference. However, existing studies often process samples independently, ignoring the spatial correlations among neighbouring objects and heterogeneity among different services. Furthermore, the inherently low interpretability of complex deep learning methods poses a significant barrier in urban planning, where transparency and extrapolability are crucial for making long-term policy decisions. To overcome these challenges, we introduce an explainable framework for inferring land use that synergises heterogeneous graph neural networks (HGNs) with Explainable AI techniques, enhancing both accuracy and explainability.  The empirical experiments demonstrate that the proposed HGNs significantly outperform baseline graph neural networks for all six land-use indicators, especially in terms of `office' and `sustenance'. As explanations, we consider feature attribution and counterfactual explanations. The analysis of feature attribution explanations shows that the symmetrical nature of the `residence' and `work’ categories predicted by the framework aligns well with the commuter's `work' and `recreation' activities in London. The analysis of the counterfactual explanations reveals that variations in node features and types are primarily responsible for the differences observed between the predicted land use distribution and the ideal mixed state. These analyses
demonstrate that the proposed HGNs can suitably support urban stakeholders in their urban planning and policy-making. The source code will be available at \url{https://github.com/xuehao0806/GNN-land-use}.


\end{abstract}




\end{frontmatter}

\section{Introduction}

Land use is a term used by urban planners to describe the function of human-made spaces in which people live, work, and recreate on a day-to-day basis. Typically, city planners are responsible for determining land use plans, shaping a city's layout by dictating what types of buildings can be located in a given space. However, due to the complex and dynamic nature of the urban environment, the actual function — both activity pattern and intensity — may deviate from the intentions of land use planning \cite{2012Inferring,2012Discovering,2020Identification}. For instance, an area initially planned as a commercial zone might gradually transition into a mixed-use area because of population growth and changes in housing demands. As the cities dynamically evolve, this deviation between land use planning and actual usage may widen, leading to inefficient use of resources, community dissatisfaction and social inequality \cite{gao2017extracting,zhang2018identifying}. To support sustainable urban planning and related policies, stakeholders need accurate and effective tools to detect the spatial distribution of human activities. 

Urban land use inference has been a challenging task for decades. Early studies primarily used simple statistical methods (e.g., linear regressions) \cite{yuan2005land, zhou2008neighborhood} and simulation-based methods (e.g., Cellular Automata) \cite{matthews2007agent, de2003stochastic} with survey data to explore urban growth patterns and dynamics with rule-based interactions. However, these traditional methods rely on over-simplified assumptions, struggling to accurately reflect complex real-world scenarios, especially for large metropolitan areas. In recent years, the advent of the era of big data has provided a new paradigm for this problem. Emergent sensor and location technologies have facilitated the collection of human activity-based data, such as mobile phone and mobility records. These passive data sources offer alternative ways for monitoring people's daily activities, facilitating the description of the urban space function. Furthermore, advanced AI models have shown significant promise in handling multi-modal and non-linear data in urban systems \cite{fadhel2024comprehensive, snavsel2024large}. These opportunities provide a strong foundation for quantitatively detecting urban land use. In the past decade, 23\% (38 out of 165) of ML-based research papers in urban spatial studies focused on inferring land use and urban form \cite{casali_machine_2022}, which shows that this problem has gradually become one of the main branches of AI applications in the field of urban studies.

Despite these recent advances, inferring land use in an accurate, explainable and thus trustworthy manner remains difficult in practice. The challenges in existing research can be summarised from 
three perspectives: data selection, methodology and post-hoc analysis.

A primary challenge lies in determining what type of activity-based data is efficient and how to fuse multi-modal data reasonably. Many activity-based datasets \cite{2012Inferring, yuan2012discovering} have been used in urban land use inference. However, inherent bias exists in many types of data when used for the representation of human activity. For instance, the number of mobile calls or messages occurring in an area could be affected by factors such as user occupation types and user communication preferences, potentially skewing observed usage intensity. Similar problems exist with social media data, which is strongly related to the user's age. Moreover, data involves user privacy, which is highly restricted for commercial use in many countries, making these types of data unavailable to many practical projects. Mobility data, on the other hand, offers low-cost, real-time, and objective insights into people’s daily movements. Meanwhile, travel demand has been systemically proven to be closely related to land use patterns \cite{bhat2009impact, ewing2010travel}. For example, a higher volume of bus passengers recorded during weekday peak hours suggests that the land use surrounding this station is work-related or residential. While mobility data can provide insights into daily human movements and associated land use patterns, a few questions remain about a) which modes of transportation data are most indicative of land use; b) how to aggregate this data across time and space effectively; and c) how to integrate data from various transportation modes to form a comprehensive analysis.

Another challenge lies in determining how to address the spatial dependency inherent in urban activities and the heterogeneity among different mobility services. Urban activities tend to cluster, exhibiting continuity along road networks and potential disruptions by physical barriers like rivers, railways and highways \cite{yigitcanlar2007gis}. Moreover, different types of mobility data contribute to different degrees of understanding surrounding land uses. Large transit hubs primarily facilitate transfers and are closely tied to adjacent transport services, whereas smaller nodes are crucial for serving `first or last mile' travel and are strongly linked to local residential and work activities \cite{2020Identification}.  Most deep learning models process samples independently, overlooking the spatial dependency between samples and their surrounding objects. Although recent studies \cite{hu2021urban,choi2022inferring} try to solve this problem by employing homogeneous graph networks, these methods often fail to encapsulate the multi-level nodes (e.g., bike stops, bus stations) and edges (e.g., road networks, tube networks) in complex urban areas. 



A third challenge lies in the fact that establishing explainability is also essential in policy and land use studies in order to build trust among government planners and policymakers. Effective land use models must be accurate, reliable, accountable for anomalies, and capable of extrapolation \cite{kleinberg2018human}. However, the black-box nature of deep learning models often hampers their trustworthiness in real-world applications. Therefore, 
a (black-box) model's credibility with planners 
needs to 
be established by systematically proving these properties through post-hoc analysis, thus reinforcing their utility in practical urban planning contexts.



Motivated by these challenges, this paper makes contributions in three key aspects: integrating mobility data, fusing multiple sources of mobility data from different travel modes using heterogeneous graph neural networks, and enhancing model explainability by using post-hoc explainable AI methods. Specifically, we make the following contributions:
\begin{itemize}
    \item We build a heterogeneous graph to integrate mobility data from multiple travel modes, including bus, tube (ie London underground) and sharing-bike, as well as topology network data, as shown in Figure~\ref{fig: example of Hetero-graph}. These rich information sources enable deep learning models to learn spatial representations that capture the heterogeneity of objects within the urban transport system. 
    \item We pioneer the application of Heterogeneous Graph Networks (HGNs) for data fusion and land use inference tasks, addressing the spatial dependency of land use features and the heterogeneity among different types of nodes and edges (Figure~\ref{fig:diagram}). Our findings indicate that the proposed method markedly enhances predictive performance over traditional neural networks and homogeneous graph neural networks.
    
    \item We pioneer the usage of feature attribution explanations in land use inference to empower explainability, addressing 
    stakeholders' concerns as to how mobility distribution in a day could affect land use predictions. We show that our model's behaviours can be verified against the stakeholders' expectations by highlighting the importance of different parts in the temporal patterns of human activity in London (UK).
    
    
    \item 
    We also define a new counterfactual explanation notion specifically suitable for explaining HGNs in land use inference tasks. We demonstrate through example analysis that our counterfactual explanations can be used to reveal changes required for the input to achieve the ideal mixed land use prediction. Through experiments, we also use such explanations to investigate the contributions of different elements in the input data to the differences between the model's predictions and the ideal mixed land use states.

\end{itemize}

The remainder of this paper is organised as follows: Section~\ref{sec:lit} presents a literature review of relevant research on land use analysis, graph deep learning, and explainable AI; Section~\ref{sec:pre} describes the data collection and pre-processing steps; Section~\ref{sec:method} outlines the methodology for Heterogeneous Graph Modelling framework and post-hoc explanations methods; Section~\ref{sec:exp} reports the results and performance metrics; Section~\ref{sec:conc} discusses the implications of our findings, limitations, and recommendations for future research.

\section{Literature Review}
\label{sec:lit}

This section will begin by reviewing the related work in land use inference. 
We then overview work on multi-modal data fusion 
and explainable AI
.

\subsection{Land use inference}
In the literature on land use, early research used simple statistical approaches (e.g., linear regressions) or simulation-based models (e.g., cellular automata) to understand urban growth patterns. Due to the development of sensor and location technologies, land use modelling has evolved from relying on survey data to incorporating activity-based big data (e.g., social media data, mobile data and transport data), enhancing understanding of urban dynamics in a fine-grained manner. In line with this change, there has been a significant shift from traditional model-driven to advanced data-driven methods. For example, Toole et al \cite{2012Inferring} applied call frequency data and correlation analysis to infer land use types in Boston, USA. The latest trend in urban land use inference incorporates big data and machine learning techniques to provide highly accurate insights and dynamic forecasts of land use. Hu et al \cite{hu2021urban} employed vehicle GPS and trajectory data, using Graph Convolutional Neural Networks, to categorise road network patterns in Beijing into commercial, public, and traffic types. Similarly, Sun et al \cite{sun2022deep} applied mobile phone data and deep convolutional autoencoders to deduce various land use types—residential, commercial, business, and industrial—in Wuhan, China.  A specific summary of activity data-based research in this domain is 
given in Table \ref{table1} below, where the first three studies use mobile data, and the subsequent ones use mobility data. 


{\small
\begin{longtable}{p{0.08\textwidth} p{0.15\textwidth} p{0.18\textwidth} p{0.20\textwidth} p{0.3\textwidth}}
\caption{Summary of papers on data-driven land use inference}\\
\toprule
\textbf{Paper} & \textbf{Scenario} & \textbf{Data} & \textbf{Method} & \textbf{Land use types} \\
\midrule
\endfirsthead

\multicolumn{5}{c}{\tablename\ \thetable: (continued)} \\
\toprule
\textbf{Paper} & \textbf{Scenarios} & \textbf{Data Type} & \textbf{Research Methods} & \textbf{Research Focus} \\
\midrule
\endhead

\bottomrule
\endlastfoot

\cite{2012Inferring} & Boston, USA & Mobile phone data & Random forests & Residential, commercial, industrial, parks\\
\cite{2012Correlating} & Harbin, China & Call frequencies & Correlation analysis & Work, residential \\
\cite{sun2022deep} & Wuhan, China & Mobile phone calls & Deep Convolutional Autoencoder & Residential, commercial, business, industrial, mixed\\
\midrule
\cite{2012Discovering} & Beijing, China & GPS Trajectory & Topic-based model & Education, residential, entertainment, mixed, tourist interest\\
\cite{2020Identification} & San Francisco, USA & Shared cycling ridership & Neural network & Residential, work, consumption, transport\\
\cite{du2020multi} & Hangzhou, China & Public bike ridership, Taxi GPS & Latent semantic classification & Residential, business, industrial, administration\\
\cite{hu2021urban} & Beijing, China & Vehicle GPS and vehicle trajectory & Graph Convolutional Neural Network & Commercial, public, traffic\\
\cite{liu2021inferring} & Beijing, China & Taxi trajectory and bus ridership & Latent clustering; Noise reduction & Residential, commercial, mixed\\
\cite{deng2022identification} & Chengdu, China & Online Car-Hailing; POI & Ensemble empirical modal decomposition & Business, office, residential, tourist interest\\
\cite{choi2022inferring} & Seoul, Korea & Taxi ridership & Statistical methods & Residential, industrial, commercial, green\\
\cite{jiao2023understanding} & Shanghai, China & Metro ridership & Spatiotemporal similarity & Residential, employment, transport, mixed\\
\cite{demissie2022understanding} & Calgary, Canada & GTFS and APC data & Statistical methods & Residential, commercial, recreational, industrial, institutional\\
\cite{zhang2023inferring} & Shenzhen, China & Surround buildings and activities data & Transformer & Residential, commercial, recreational, industrial, institutional\\
\label{table1}
\end{longtable}
}


A recent review paper found that 23\% (38 out of 165) of AI-based research papers in urban spatial studies focused on inferring land use and urban form \cite{casali_machine_2022}, which showed that inferring land use has become one of the main branches of AI application in the field of city planning domain. However, this field is still in its early developmental stages, and in practice faces numerous challenges in maintaining accuracy and trustworthiness.

One major gap lies in fusing multi-modal data to overcome the potential bias and incompleteness of each data source. Given Table \ref{table1}, two forms of mobility data typically used are: trajectory data and ridership data (inbound and outbound flow at the station level). Trajectory data reflects usage intensity at a road level but fails to accurately represent the land use function of surrounding space. For example, a highway from a suburban area to a city centre might predominantly serve daily commuters, but this does not necessarily mean there are significant residential or work facilities along this highway. On the other hand, ridership data, which measures the flow of passengers into and out of stations, provides a more intuitive reflection of travel related to human activities. However, a single type of transit facility is typically not able to cover the entire urban space.  Data fusion can be an effective way to address the lack of spatial coverage and potential biases. Yet, only a few studies \cite{liu2021inferring} have performed data fusion (e.g., ensemble methods at the output level) to merge different datasets, which limits the exploration of intrinsic spatial heterogeneity and complementary relationships within the mobility system.

Another gap exists in the methods for improving the explainability of land use inference models. The successful application of AI in critical domains largely depends on whether the model is explainable and trustworthy. However, explainability often represents a trade-off with accuracy \cite{Ali_23} and trustworthiness has been shown to be a complex phenomenon which is affected by explainability in many ways \cite{Jacovi_21,Schoeffer_22,Ferrario_22,Zerilli_22,Panigutti_22}. Indeed, the black-box nature of complex data-driven models often hampers their adoption in real-world applications, particularly in urban planning and policy-making where transparency and accountability are highly valued \cite{papadakis2024explainable}. Besides, stakeholders anticipate that the model can go beyond the prediction of the outcome; they also expect it to be able to respond to hypothetical or ``what-if" scenarios, like ``what efforts can be made to achieve sustainable land use planning". However, few land use inference studies have provided explainable modelling or post-hoc explanations for their proposed models.

\subsection{Multi-modal data fusion}


The need for fusion of multi-modal data for better accuracy and robustness is not unique to land use inference but also exists in many efforts for the development of smart and sustainable cities, such as traffic prediction \cite{jin2021hetgat,li2024towards}, energy management \cite{chui2021handling, wang2021energy}, and healthcare \cite{schwalbe2020artificial}. In urban studies, commonly used fusion strategies can be divided into the following categories: fuzzy logic-based \cite{snavsel2024large, zheng2021fusion}, Bayesian-based \cite{du2022bayesian, wang2020bayesian}, ensemble-based \cite{ganaie2022ensemble, ren2016ensemble}, and graph-based \cite{snavsel2024large, liang2024time, zhou2022identifying}. For the mobility data discussed in this paper, graph-based methods like GraphSages \cite{hamilton2017inductive} and Graph Attention Networks \cite{velivckovic2017graph} are particularly effective due to the strong ability to process spatial dependency among the facilities in a large-scale graph form. For example, Hu et al \cite{hu2021urban} introduced a Graph Convolutional Network (GCN) to understand the social function of road segments, showing significant accuracy compared with traditional machine learning.

To better characterise the spatial heterogeneity and hierarchy inside the mobility system, we propose a novel hetero-graph deep learning framework for mobility data fusion. As an extended version of Graph Neural Networks (GNNs) – Heterogeneous Graph Networks (HGNs) are designed to handle graph structures with multiple types of nodes and edges. Existing studies have proven that HGNs are highly effective in fusing multi-modal data and representing complex graph structures in various domains, such as recommendation systems \cite{shi2018heterogeneous}, biomedical structures \cite{wu2023survey}, and social networks \cite{wang2022survey}. In scenarios like urban systems, HGNs have been applied to critical problems like traffic prediction, and signal control with significant performance improvement \citep{li2024towards, yang2022inductive}. However, no research has been undertaken so far on techniques related to the use of HGNs in land use inference. This paper proposes a novel framework using an attention-based heterogeneous graph neural network model to effectively represent the complex interactions within multi-modal mobility systems.

\subsection{Explainability}

Recent works on graph-based explanation have provided new tools for understanding the function of elements in heterogeneous mobility graphs. Given the requirement of this paper, two types of analytic XAI methods are considered: feature attribution analysis and counterfactual analysis.
\begin{itemize}
    \item Feature attribution methods quantify the approximate importance of 
    individual input features on a particular model prediction through an attribution value. These techniques, such as Integrated Gradients \cite{sundararajan2017axiomatic} and SHapley Additive exPlanations (SHAP) \cite{lundberg2017unified}, clarify the roles of input features in the prediction of the outcome and have been successfully used in understanding graph elements (nodes and edges) in many critical domains, such as molecular science 
    \cite{yan2020graph, bi2020interpretable} and Internet of Things \cite{dobrojevic2023addressing}.
    \item Counterfactual explanations 
     demonstrate 
     which changes are needed in the factual input to achieve a different output prediction \cite{Wachter17}.  
     These methods have been used in domains like molecular science \cite{numeroso2021meg} and recommendation systems \cite{chen2022grease} to understand how potential changes in the graph might result in desired outcomes. 
\end{itemize}

Although these XAI methods have been used to explain GNNs in many critical domains, no studies have applied them to the land use inference problem. Facing this gap, this paper pioneers the use of the integrated-gradient method to analyse how mobility distribution influences the inferred land use types. Additionally, we introduce a new counterfactual explanation notion designed specifically for heterogeneous graph structures, aimed at identifying necessary minimum input modifications to achieve the ideal mixed land use state.


\section{Preliminaries}
\label{sec:pre}
This section introduces the research problem and the basic concepts for building heterogeneous graph models. The notation used throughout this paper is shown in Table \ref{table: notation}.

\begin{table}[ht!]
\centering
\caption{Notation and Descriptions. The table is divided into three sections: the first section includes general operations; the second section includes components of the graph; and the third section includes elements related to the learning problem. 
}
\begin{tabular}{>{\bfseries}l p{11.5cm}}
\toprule
\textbf{Notation} & \textbf{Description} \\ \midrule
\( \oplus \) & A concatenation operation \\
\( \lvert * \rvert \) & The cardinality of a set \\
\midrule
\(\mathcal{G}\) & A graph \\
\(\mathcal{V}\) & A set of nodes\\
\(\mathcal{E}\) & A set of edges\\
\(\mathcal{A}\) & A set of node types \\
\(\mathcal{R}\) & A set of edge types \\
\(v_i \in \mathcal{V} \) & A single node with subscript $i$ (also referred to simply as $i$)\\
\(e_{ij}\in \mathcal{E}\) & An edge representing a pair of nodes \(( v_i, v_j)\) iff $v_i$ and $v_j$ are connected\\
\(\phi\) & A node-to-type mapping \\
\(\psi\) & An edge-to-type mapping \\
\midrule
\(\mathcal{F}\) & A set of features of each node 
\\
\(\mathcal{T}\) & A set of land use targets 
\\
\( x_{i, j} \) & The $j$-th feature for node \( i \) \\
\( h^{(0)}_i \) & The initial node representation after the feature transformation \\
\( h^{(l)}_s \) & The updated representation of target node $s$ at layer \( l \) \\
$\Theta$ & The set of parameters for training \\
\( W_{\phi(i)}^{\text{P}}\) & The initial projection matrix for node type \( \phi(i) \) \\
\( W_{\tau,\phi(s)}^{\text{MH}} \) & The multi-head matrix for source node type in head $\tau$\\
\( W_{\psi(e)}^{\text{MP}} \) & The message passing matrix for edge type $\psi(e)$ \\
\( W_\tau^\text{Q}, W_\tau^\text{K},  W_\tau^\text{A} \) & The query, key and attention matrices for head \( \tau \) \\
\( W^{\text{AG}}_{\phi(t)} \) & The aggregation matrix for target node type $\phi(t)$ \\
\( m^{(l)}(s, e, t) \) & The message representation from \( v_s \) to \( v_t \) through edge $e$ at layer \( l \) \\
\( \alpha^{(l)}(s, e, t) \) & The attention weights from \( v_s \) to \( v_t \) through edge $e$ at layer \( l \) \\
\( \xi \) & The scaling factor in the attention mechanism \\
\( \theta \) & The number of heads in the multi-head mechanism \\
\bottomrule
\end{tabular}
\label{table: notation}
\end{table}

We begin by defining heterogeneous graphs. 

\begin{definition} \textbf{Graph:}
A graph \( \mathcal{G} \) is represented as \((\mathcal{V}, \mathcal{E})\),
where \( \mathcal{V} \) is a set of nodes (or vertices), i.e., \( \mathcal{V} = \{v_1, \ldots, v_{|\mathcal{V}|}\} \), and \( \mathcal{E} \) denotes a set of edges connecting nodes, i.e., \( \mathcal{E} = \{e_1, \ldots, e_{|\mathcal{E}|}\} \). Each edge \( e \in \mathcal{E} \) is a pair of nodes \( (v_i, v_j) \) iff 
$v_i$ and $v_j$ are connected. 

\end{definition}
In a heterogeneous graph nodes and/or edges belong to different types, as follows:

\begin{definition} \textbf{Heterogeneous graph:}
Let \( \mathcal{A} \) denote a set of node types and  \( \mathcal{R} \) denote a set of edge types, such that \( |\mathcal{A}| + |\mathcal{R}| > 2 \).
Then a heterogeneous graph \( \mathcal{G}\) is \((\mathcal{V}, \mathcal{E})\) such that there exists a function \( \phi: \mathcal{V} \rightarrow \mathcal{A} \) that maps each node with its corresponding node type and a function \( \psi: \mathcal{E} \rightarrow \mathcal{R} \) that maps each edge with its corresponding edge type.
\end{definition}
\begin{definition}\textbf{Meta relation:}
The meta relation for an edge $e = (v_s, v_t)$ linked from source node $v_s$ to
target node $v_t$ 
is denoted as $\langle \phi(s), \psi(e), \phi(t) \rangle$. 

\end{definition}

\begin{figure}
    \centering
    \includegraphics[width=1.0\linewidth]{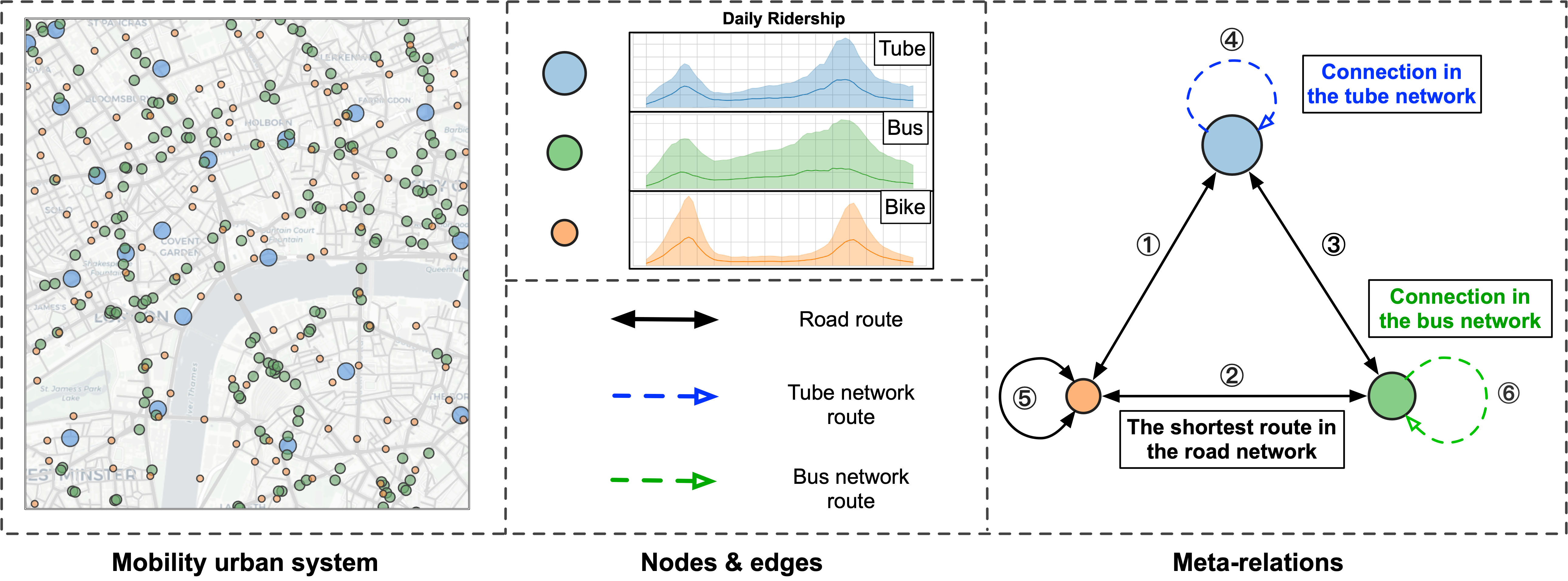}
    \caption{This figure depicts a heterogeneous mobility network featuring meta-relations that highlight various connection types. The left panel displays a central London area with nodes coloured by different types of mobility services, representing tube stations (blue), bus stops (green), and bike stations (orange). The middle panel presents a legend for road routes (solid black lines), tube routes (dashed blue lines), and bus routes (dashed green lines), accompanied by daily ridership graphs for each mode of transport
    . The right panel outlines meta-relations with nodes and their connections, illustrating the complexity of information in a multi-modal mobility network.}
    \label{fig: example of Hetero-graph}
\end{figure}

Urban mobility systems can be conceptualised as heterogeneous graph structures with multiple types of service facilities in the form of nodes (e.g., tube and bus stations) and spatial connections among them in the form of edges (e.g., transit routes). Specifically, we consider two categories of edges: edges in the road network and edges in the transit networks. When travellers transfer between different types of stations, they usually need to walk. Therefore, as shown in Figure \ref{fig: example of Hetero-graph}, we regard the connections across various types of nodes as the shortest path on the road network (\textcircled{1}, \textcircled{2}, \textcircled{3}). For edges between nodes of the same type, there may be shortest routes in a road network (\textcircled{4}) or connecting paths in a specific network, such as a tube network or a bus network (\textcircled{5} and \textcircled{6}).


Given the complexities 
in the data, this paper aims to build an explainable heterogeneous graph-based framework for the land use inference problem. To begin with, the research is framed as a multi-task regression problem. This setting stems from practical considerations: compared with merely categorising land use types, it is more useful for stakeholders to obtain the intensities of different land use because of the need to analyse diversity \cite{2020Identification}.

Mathematically,  we build a multivariate mapping function \(\mathcal{M}\) that regresses land use intensity based on daily mobility distribution and urban heterogeneous graph structure, shown as follows:

\begin{equation}
Y=\mathcal{M}_{\Theta}\left(\mathcal{G} ;X\right)
\end{equation}
 where the parameters set for training are $\Theta$, the nodes feature $X \in \mathbb{R}^{| \mathcal{V}| \times |\mathcal{F}|}$ ($\mathcal{F}$ is the set of features of each node\footnote{
 Concretely, we denote nodes features as the daily passenger inbound data of each station (15-minute interval ridership from 6:00 to 22:00, forming 64 input features).}) represents the mobility inbound or outbound variables at different time intervals, such as `15-min passengers inflow in a typical day'\footnote{A `typical day' refers to the mean values of data collected over multiple statistical periods (weeks, months) under normal, non-exceptional conditions \cite{2020Identification}.} for all nodes in $\mathcal{G}$. $Y \in \mathbb{R}^{| \mathcal{V}| \times |\mathcal{T}|}$ ($\mathcal{T}$ is the set of targets\footnote{Concretely, we use, as $\mathcal{T}$, the land use indicators in this task (e.g., office and residence).}) denotes the multi-regression targets. 
The goal of land use inference is to find an optimised set of \(\Theta^{*}\) to minimise the loss between the estimated land use indicators and land use labels:

 \begin{equation}
     \Theta^*=\underset{\Theta}{\arg \min } \frac{1}{n} \sum_{i=1}^n \operatorname{loss}(\mathcal{F}_{\Theta}\left(\mathcal{G} ;X\right), \hat{Y})
 \end{equation}
where \(\hat{Y}\in \mathbb{R}^{| \mathcal{V}| \times |\mathcal{T}|}\) refers to the land use ground-truth labels 
and $n$ refers to the number of samples in the training dataset.

\section{Methodology}
\label{sec:method}

\begin{figure}
    \centering\includegraphics[width=1.0\linewidth]{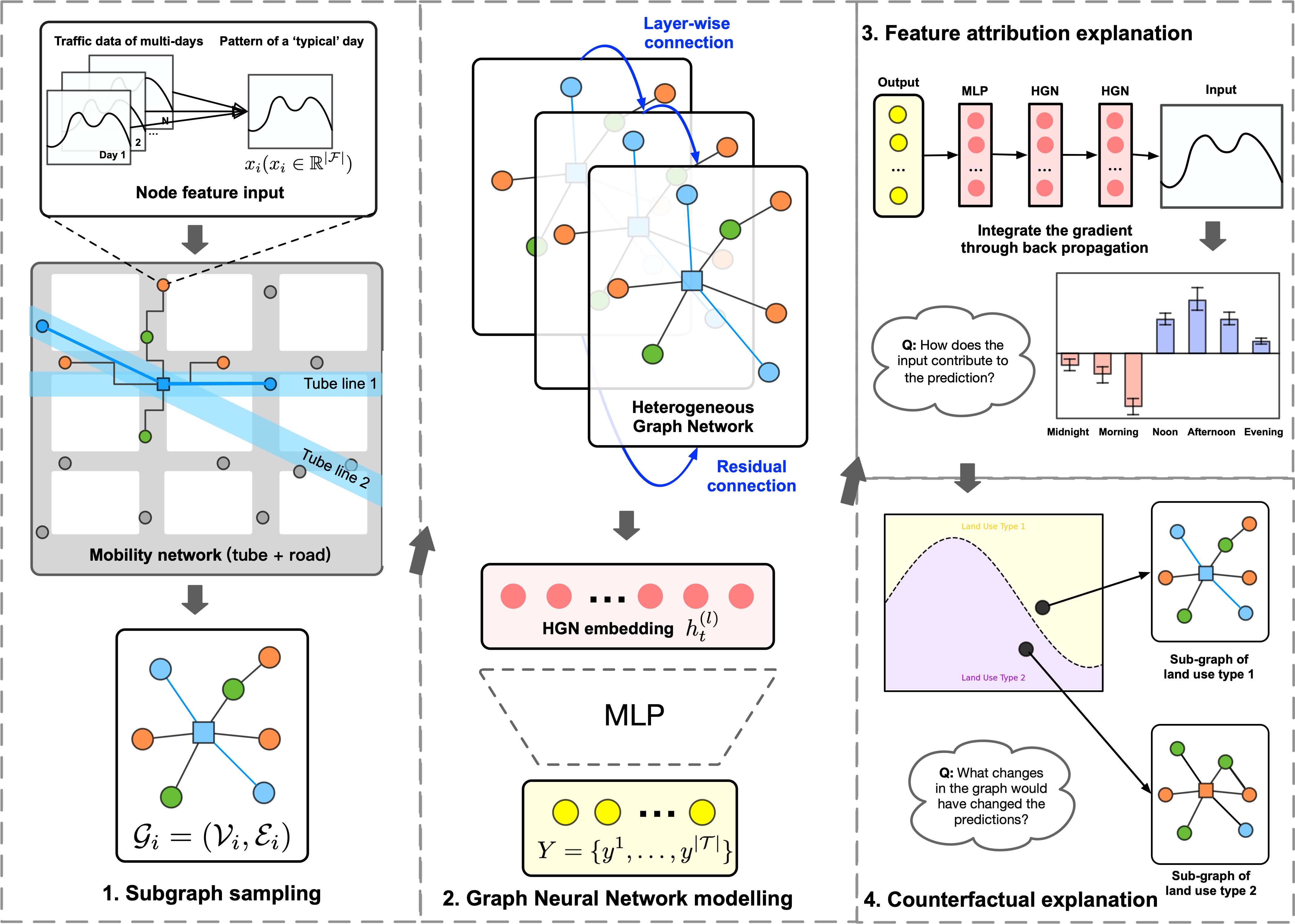}
    \caption{
    Overview of 
    our framework for land use inference based on multi-mobility systems.}
    \label{fig:diagram}
\end{figure}

Figure \ref{fig:diagram} illustrates the framework of this methodology, divided into three main parts: a) Graph Building and Sampling: synthesising traffic data into typical daily patterns, collecting edge information from a hierarchical urban mobility network, and sampling heterogeneous subgraphs; b) Graph Neural Network Modelling: processing these subgraphs through a HGN with layer-wise and residual connections, embedding the output into a feature vector, and using a fully connected linear layer to predict land use targets; c) Explanation Techniques: using feature attribution explanations to determine the contribution of input features to the predictions and counterfactual explanations to explore the impact of changes in the graph structure on prediction outcomes.


\subsection{Heterogeneous Attention Embedding}
Given the rich mobility information and road topology information, our primary focus is on effectively embedding structural data and node-level mobility features within urban mobility networks. Inspired by the recent progress of HGNs in social network analysis \cite{jiang2023meta}, we propose a Heterogeneous Graph Encoder based on HGNs. 
The HGN module aims to learn node representations through predetermined meta-relations. To model the influence of various node types, we utilise a multi-head attention mechanism.


Firstly, we define a node feature transformation function that aligns features \( x_i \) of nodes $v_i$ to a unified feature space, based on their node types \( \phi(i) \). This is formulated as: 
\begin{equation}
h^{(0)}_i = W_{\phi(i)}^{\text{P}} \cdot x_i,
\end{equation}
where \( W^\text{P} \) is a type-specific linear projection matrix. Through this, we derive the initial node representation \( h^{(0)}_i \) for node \( i \).

Subsequently, we implement a multi-head mechanism to compute the message representation \( m^{(l)}(s, e, t) \) at layer \( l \geq 0\) from source nodes \( v_s \) to target nodes \( v_t \). This process accommodates the types of nodes \( \phi(s), \phi(t) \)  and edges \( \psi{(e)} \) 
in the heterogeneous graph, as expressed:

\begin{equation}
m^{(l)}(s, e, t)=\underset{\tau \in[1, \theta]}{\oplus}\left(\left(W_{\tau,\phi(s)}^{\text{MH}} h_s^{(l-1)}\right) W_{\psi(e)}^{\text{MP}}\right)
\end{equation}
Here, \( \oplus \) represents the concatenation operation, and \( \theta \) the total number of heads, while \( W_{\tau,\phi(s)}^{\text{MH}} \) and \( W_{\psi(e)}^{\text{MP}} \) are matrices designated for processing based on node and edge types.

Next, we apply a multi-head attention framework to derive attention weights \( \alpha^{(l)}(s, e, t) \), facilitating the interaction between source \( v_s \) and target \( v_t \) nodes via edges \( \psi(e) \), described as follows:

\begin{equation}
\text{head}^{(l)}_\tau (s, e, t) = \left( \left( W_\tau^\text{Q} h^{(l-1)}_s \right)^T W_\tau^\text{A} \left( W_\tau^\text{K} h^{(l-1)}_t \right)^T \right) \frac{1}{\sqrt{\xi}},
\end{equation}

\begin{equation}
\alpha^{(l)}(s, e, t) = \text{Softmax} \left( \underset{\tau \in[1, \theta]}{\oplus} \text{head}^{(l)}_\tau (s, e, t) \right),
\end{equation}
where \( W_\tau^\text{Q}, W_\tau^\text{K} \), and \( W_\tau^\text{A} \) are the linear projection matrices per head \( \tau \), tailored according to node and edge types, and \( \xi \) is an adaptive scaling factor for the attention mechanism.

Finally, the information is aggregated from the neighbours of the target node with attention weights to update the representation of the target node, expressed as:
\begin{equation}
h^{(l)}_t = W^{\text{AG}}_{\phi(t)} \left( \sum_{v_s \in \mathcal{N}(v_t)} \left( \alpha^{(l)}(s, e, t) \cdot m^{(l)}(s, e, t) \right) \right) + h^{(l-1)}_t,
\end{equation}
with \( W^{\text{AG}} \) as the aggregation-specific linear matrix considering the type of the target node. Here, \( \mathcal{N}(v_t) \) denotes the set of neighbour nodes of the target node \( v_t \).
\subsection{Explainable AI methods for post-hoc explanations}

This section discusses the XAI methods to enhance explainability for land use inference: feature attribution analysis and counterfactual analysis. Two main reasons drive this need: stakeholders, particularly in urban planning and development, need to understand the importance (to decision-making) of specific elements in predictive models, especially for intricate input-outcome relationships. This justifies the use of feature attribution explanations. Additionally, these stakeholders seek to identify minimal changes that can be made to achieve desired outcomes, such as a community with balanced land use. This is our reason for choosing to also include counterfactual explanations. Details of these two methods are shown follow.

\subsubsection{Feature attribution methods}
The first class of techniques we use to explain the predictions of our model are \emph{feature attribution methods}. These techniques quantify the approximate importance of the individual input features on a particular model prediction through linear \emph{feature attribution scores}. Such scores can provide invaluable insights regarding the reasoning of deep learning models and uncover possible issues or biases \cite{pezeshkpour-attribution-artifacts, meng-interpretability-fairness-mimic}.


In the context of explaining an HGN model $\mathcal{M}$ applied to a heterogeneous graph \( \mathcal{G} = (\mathcal{V}, \mathcal{E}) \), we can formalise the feature attribution scores as follows. Given a target node in the graph $v_i \in \mathcal{V}$ and a target output $y_{i, j}$, our goal is to compute the attribution scores $S \in \mathbb{R}^{|\mathcal{V}| \times |\mathcal{F}|}$ estimating the importance of every node and feature combination for the model output $y_{i, j}$. Intuitively, the element $S_{k, l}$ of the attribution score matrix provides an estimate of how much the $l$-th feature of the $k$-th node, $x_{k, l}$, affected the $j$-th output at the $i$-th node, $y_{i, j}$. To get a more holistic view of the importance of the different features, we will also consider the aggregated attribution scores $a \in \mathbb{R}^{|\mathcal{F}|}$ where $a_k = \sum^{|\mathcal{V}|}_{i = 0} S_{i, k}$. Each element of the aggregated attribution scores vector estimates how much a particular input feature affected the target output when considering its combined effect from all the nodes in the graph.

In our work, we use two different feature attribution methods for explaining our HGN model — Gradient $\cdot$ Input \cite{shrikumar-old-deeplift-gxi} and Integrated Gradients \cite{sundararajan-integrated-gradients}. Gradient $\cdot$ Input is a simple gradient-based attribution method defined as follows, in our setting:

\begin{definition}
    \textbf{Gradient $\cdot$ Input Explanation}. Given an explained model $\mathcal{M}$, an input graph $\mathcal{G} = (\mathcal{V}, \mathcal{E})$, a target node $v_i \in \mathcal{V}$ and a target output $y_{i, j}$, the Gradient $\cdot$ Input feature attributions scores are defined as:
    \begin{align*}
        S^{G \cdot I}_{k, l} = x_{k, l} \cdot \frac{\partial y_{i, j}}{\partial x_{k, l}} = x_{k, l} \cdot \frac{\partial \mathcal{M}(\mathcal{G})_{i, j}}{\partial x_{k, l}}
    \end{align*}
\end{definition}

The advantage of Gradient $\cdot$ Input is its computational efficiency, its conceptual simplicity as well as its good ability to faithfully capture the behaviour of the model in the small region around the considered input \cite{ancona-towards-attribution}. However, the attributions produced by Gradient $\cdot$ Input may also be unreliable, particularly if the function of the considered model is highly irregular (resulting in a problem known as shattered gradients \cite{balduzzi-shattered}) or saturates for the given input (e.g., due to the predicted output already being sufficiently high). These limitations are largely addressed by Integrated Gradients explanations, which are defined as follows, in our setting:

\begin{definition}
    \textbf{Integrated Gradients Explanation}. Given an explained model $\mathcal{M}$, an input heterogeneous graph $\mathcal{G} = ( \mathcal{V}, \mathcal{E} )$, a target node $v_i \in \mathcal{V}$, a target output $y_{i, j}$ and a reference graph $\overline{\mathcal{G}} = ( \overline{\mathcal{V}}, \overline{\mathcal{E}} )$, the Integrated Gradients feature attributions scores are defined as:
    \begin{align*}
        S^{IG}_{k, l} = (x_{k, l} - \overline{x_{k, l}}) \cdot \int^{1}_{\alpha = 0} \frac{\partial \mathcal{M}(\overline{\mathcal{G}} + \alpha \cdot (\mathcal{G} - \overline{\mathcal{G}})_{i, j}}{\partial x_{k, l}} d\alpha
    \end{align*}
\end{definition}

In contrast with Gradient $\cdot$ Input, which only considers the gradient at the explained input instance, Integrated Gradients are based on an aggregate gradient on the path from some suitable reference to the currently considered input. This helps to naturally smooth out any local irregularities in the gradient and to capture the influence of the input features even if their effect already saturated close to the currently considered input \cite{sundararajan-integrated-gradients}. In our experiments, we will consider the neutral reference graph $\overline{\mathcal{G}}$ with the same structure as the input graph $\mathcal{G}$, but all the features on the nodes and edges set to zero. Using zero feature values has been advocated as a reasonable default choice for the reference input \cite{sundararajan-integrated-gradients}. We must also note that Integrated Gradients in the pure form defined above are typically intractable for complex deep learning models. Thus, in practice, the integrated gradient scores are computed, in our setting, via the following approximation:

\begin{definition}
    \textbf{Approximate Integrated Gradients Explanation}. Given an explained model $\mathcal{M}$, an input heterogeneous graph $\mathcal{G} = ( \mathcal{V}, \mathcal{E} )$, a target node $v_i \in \mathcal{V}$, a target output $y_{i, j}$, a reference graph $\overline{\mathcal{G}} = ( \overline{\mathcal{V}}, \overline{\mathcal{E}} )$ and a number of steps $n$, the approximate Integrated Gradients feature attributions scores are defined as:
    \begin{align*}
        S^{IG-A}_{k, l} = (x_{k, l} - \overline{x_{k, l}}) \cdot \sum^{n}_{k = 0} \frac{1}{n} \cdot \frac{\partial \mathcal{M}(\overline{\mathcal{G}} + \frac{k}{n} \cdot (\mathcal{G} - \overline{\mathcal{G}})_{i, j}}{\partial x_{k, l}}
    \end{align*}
\end{definition}

Throughout this paper, when considering Integrated gradient explanations, we will use the above approximation with $n = 50$ as the number of steps.

In addition to the above gradient-based methods, we also initially considered perturbation approaches based on Shapley Values \cite{lundberg2017unified}. However, given the large size of the considered input graph, we found these to be highly computationally inefficient, taking several hours to compute the explanation for a single output. We also note that there are several bespoke methods specifically focused on explaining GNN models, such as GNN-Explainer \cite{gnnexplainer-ying}. Unfortunately, the available implementations for these methods were not compatible with the library for HGNs that we used, which is why we opted to use the above methods instead. Extending GNN-Explainer and other GNN-specific explainability methods to our heterogeneous model poses an interesting line of future work, but is outside the scope of this paper. 

\subsubsection{Counterfactual Explanations}


In contrast to attributing model output predictions to parts of input features, Counterfactual Explanations (CEs) reveal the minimal changes needed from the factual input to achieve a different and more desirable output prediction \cite{Wachter17}. 
In this section, we define CEs suitable for the land use inference task with heterogeneous graphs. 
For graph data, CEs often take the form of subgraphs or modified graph inputs \cite{DBLP:journals/csur/PradoRomeroPSG24} due to the connectivity between the nodes, which is especially the case in land use inference since our interest is always on an area \cite{casali2022machine}. We start by defining the subgraph consisting of a target node and its neighbouring nodes.

\begin{definition}
    \textbf{1-hop complete subgraph}. Given an input heterogeneous graph $\mathcal{G}=(\mathcal{V}, \mathcal{E})$ and a target node $v_i \in \mathcal{V}$, the \emph{1-hop complete subgraph} of $v_i$ is defined as $\mathcal{G}_i=(\mathcal{V}_i, \mathcal{E}_i)$, where $\mathcal{V}_i = \{v_j \text{ } | \text{ } e_{i,j} \in \mathcal{E} \vee e_{j,i} \in \mathcal{E}\} \cup \{v_i\}$ and $\mathcal{E}_i = \{e_{j,k} \text{ } | \text{ } v_j, v_k \in \mathcal{V}_i, \text{ } e_{j,k} \in \mathcal{E}\}$.
\end{definition}

Another concept central to CEs for graph data is the dissimilarity measure between two subgraphs, which allows comparisons of candidate CEs. Specifically, suppose we have two candidate subgraph CEs; that which is more similar to the input is considered the better candidate, because it would require less effort to change from the input \cite{Wachter17}. Given the multi-modal nature of our data, we design a dissimilarity measure covering the differences in node features, node types, edge types, and (sub)graph structure as follows:

\begin{definition}
    \textbf{Local subgraph dissimilarity}. Given an input heterogeneous graph $\mathcal{G}=(\mathcal{V}, \mathcal{E})$, two 1-hop complete subgraphs $\mathcal{G}_i=(\mathcal{V}_i, \mathcal{E}_i)$ and $\mathcal{G}_j=(\mathcal{V}_j, \mathcal{E}_j)$ for nodes $v_i, v_j \in \mathcal{V}$, respectively, the \emph{local subgraph dissimilarity} between $\mathcal{G}_i$ and $\mathcal{G}_j$ is:

    \[
        d(\mathcal{G}_i, \mathcal{G}_j) = d_{\mathcal{V}}(\mathcal{G}_i, \mathcal{G}_j) + d_{\mathcal{A}}(\mathcal{G}_i, \mathcal{G}_j) + d_{\mathcal{R}}(\mathcal{G}_i, \mathcal{G}_j) + d_{\mathcal{G}}(\mathcal{G}_i, \mathcal{G}_j)
        \]
    

where:     
\begin{align*}
d_{\mathcal{V}}(\mathcal{G}_i, \mathcal{G}_j) & = \frac{1}{\lvert \mathcal{V}_j \rvert} \sum_{v_k \in \mathcal{V}_j} \frac{1}{2}(\textit{L2\_dist}(x_{i}, x_{k}) + \textit{cosine\_dist}(x_{i}, x_{k})) 
\end{align*}
is the \emph{node feature dissimilarity}, \textit{L2\_dist} and \textit{cosine\_dist} are L2 (Euclidean) distance and cosine distance respectively,
\begin{align*}
d_{\mathcal{A}}(\mathcal{G}_i, \mathcal{G}_j) & = 1 - \frac{\sum_{k=1}^{\lvert \mathcal{A} \rvert} min(\overline{\phi(\mathcal{V}_i)}_k), \overline{\phi(\mathcal{V}_j)}_k)}{\sum_{k=1}^{\lvert \mathcal{A} \rvert} max(\overline{\phi(\mathcal{V}_i)}_k), \overline{\phi(\mathcal{V}_j)}_k)} 
\end{align*}
is the \emph{node type dissimilarity}, $\phi$ maps nodes to its corresponding type, $\phi(\mathcal{V}_i)$ is the node types of subgraph nodes $\mathcal{V}_i$, $\overline{\phi(\mathcal{V}_i)} \in \mathbb{R}^{\lvert \mathcal{A} \rvert}$ represents the counts of each node type in $\mathcal{A}$, 
\begin{align*}
d_{\mathcal{R}}(\mathcal{G}_i, \mathcal{G}_j) & = abs(\frac{1}{\lvert \mathcal{E}_i \rvert} \sum_{e_{k,l} \in \mathcal{E}_i} \psi({e_{k,l}}) - \frac{1}{\lvert \mathcal{E}_j \rvert} \sum_{e_{m,n} \in \mathcal{E}_j} \psi({e_{m,n}})) 
\end{align*}
is the \emph{edge type dissimilarity}, $\psi$ maps edges to its corresponding type,
\begin{align*}
d_{\mathcal{G}}(\mathcal{G}_i, \mathcal{G}_j) & = |\lvert \mathcal{E}_i \rvert - \lvert \mathcal{E}_j \rvert|
\end{align*}
is the \emph{graph structure dissimilarity}, $abs(*)$ takes the absolute value of a real number.

\end{definition}



This score combines 
four sources of dissimilarities. Firstly, $d_{\mathcal{V}}$ denotes the node features differences 
where L2 captures the magnitude differences, while cosine distance captures the traffic pattern differences. $d_{\mathcal{A}}$ calculates the node type differences between the nodes in the two subgraphs using the multiset Jaccard distance \cite{costa2021further}. $d_\mathcal{R}$ calculates the absolute differences between the average road types, obtained by the edge types in both subgraphs. Finally, $d_\mathcal{G}$ represents the graph structure differences expressed as the absolute difference between the numbers of edges, which is often considered as sparsity in graph CEs \cite{DBLP:journals/csur/PradoRomeroPSG24}. We now use Local subgraph dissimilarity as a core heuristic\footnote{Note that our dissimilarity notion is not a distance, e.g., it does not satisfy symmetry, as $d(\mathcal{G}_i, \mathcal{G}_j)\neq d(\mathcal{G}_j, \mathcal{G}_i)$.} to guide the search for CEs:


\begin{definition}
    \textbf{Counterfactual Explanation}. Given an explained model $\mathcal{M}$, an input heterogeneous graph $\mathcal{G}=(\mathcal{V}, \mathcal{E})$, a target node $v_i \in \mathcal{V}$ with predicted output $y_{i}$, a desirable prediction result $y'\neq y_{i}$, the \emph{counterfactual explanation} for $v_i$'s prediction is $\mathcal{G}_{j}=(\mathcal{V}_j, \mathcal{E}_j)$, the 1-hop complete subgraph for node $v_j$, which is:
    \[
    \underset{v_j \in \mathcal{V} |  y_j = y'}{\textit{argmin}} \quad d(\mathcal{G}_i, \mathcal{G}_j)
    \]
\end{definition}


Our definition finds a 1-hop complete subgraph from the given heterogeneous graph data as the CE for a given target node. It is the closest (measured by local subgraph dissimilarity) subgraph to the 1-hop complete subgraph of the target node, among all the subgraphs of nodes which are predicted with a pre-specified desirable prediction by our explained model. Our method has the same spirit as the nearest-neighbour CEs method for tabular data, which finds the closest point in the training dataset as the CE for an input point \cite{NiceNNCE}. Such CEs taken from the dataset have the advantage of lying within the data manifold, therefore they are realistic data and not algorithmic artefacts \cite{DBLP:conf/aies/PoyiadziSSBF20}. Additionally, they are more robust to perturbations that could occur in the inference pipeline \cite{DBLP:journals/corr/jiangsurvey24}.

Practically, the CEs can be computed as the following. First, initialise the 1-hop complete subgraph for the target node. Then, filter out the nodes predicted with the desirable label. In an iterative procedure, for each of these nodes, take their 1-hop complete subgraph, and compute the local subgraph dissimilarity between the subgraph and the input node's subgraph. The subgraph with the lowest dissimilarity is returned as the CE. As will be seen in Section \ref{ssec:counterfactual_analysis}, we specify a \textit{mixed} prediction and use CEs to investigate what changes are needed to achieve a community with balanced land use \cite{farr2011sustainable, lehmann2016sustainable}. 


\section{Empirical Experiments}
\label{sec:exp}

\subsection{Dataset description}

\begin{figure}[htbp]
    \centering
    \includegraphics[width=1\linewidth]{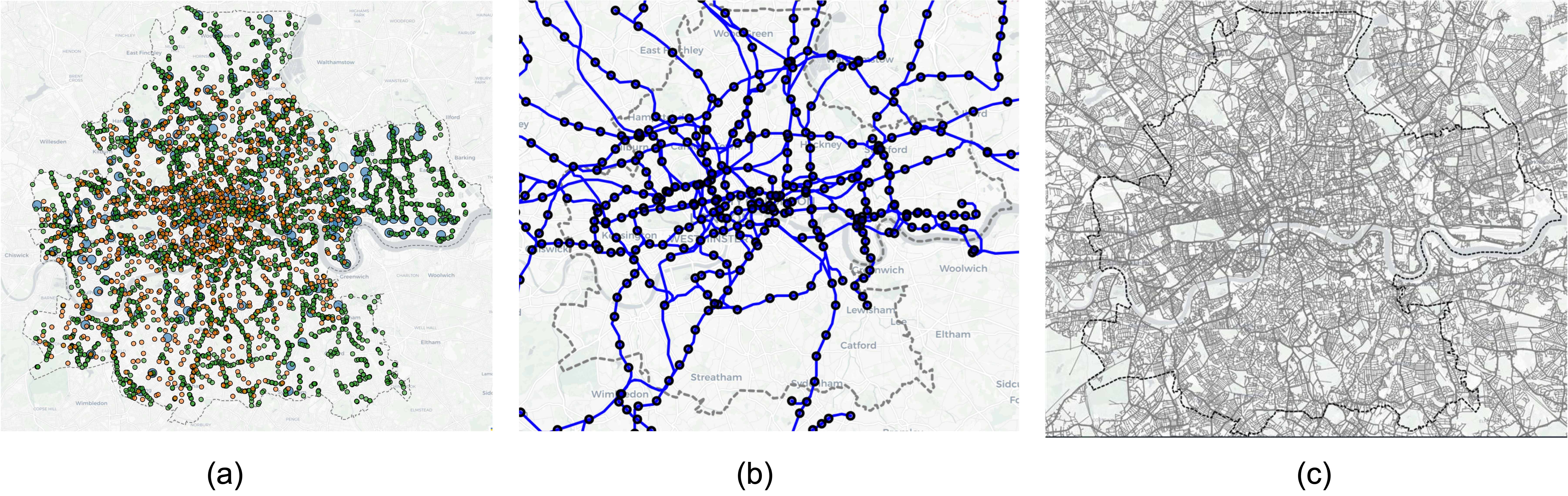}
    \caption{The heterogeneous input is based on three data sources: a) 
    locations of the tube, bus, and cycling stations, marked in blue, green, and orange respectively; b) 
    connectivity among 
    tube stations in the tube network; c) 
    road network at different levels, which is used to build neighbourhood connections.}
    \label{fig: input demonstrations}
\end{figure}

Since no open-source database currently offers well-processed graph datasets concerning urban mobility data and transport facilities, the first step of our case study is to construct a heterogeneous graph dataset of multi-modal urban mobility information. As shown in Figure~\ref{fig: input demonstrations}, London (inner and central) is chosen as the study area due to the high complexity and diversity of its land use. In addition, its dense mobility network allows for fine-grained regional analysis to cover most of the study area.

 The constructed dataset comprises four parts: node features, edge features, labels and zoning as described below.

\begin{itemize}

    \item \textbf{Nodes features}. For any mobility station, we use its ridership distribution on a `typical weekday'(as averaged over the weekdays in 2017) as the input nodes feature. This ridership information is provided by the Transport of London (TfL) open data  (
    \url{tfl.gov.uk/info-for/open-data-users/our-open-data}). Spe\-cifically, three types of mobility services are selected -- tube (186 stations), bus (3167 stations) and bike sharing (883 stations) 
    -- as shown in Figure \ref{fig: input demonstrations}(a). The daily passenger inbound data of each station is processed into 15-minute interval ridership from 6:00 to 22:00 (64 input features). 
    \item \textbf{Edge features}. The edges are of two types: tube network edges, derived from connectivity data among tube stations, as shown in Figure \ref{fig: input demonstrations}(b), and road network edges, based on a comprehensive road network, shown in Figure \ref{fig: input demonstrations}(c). There are specifically five types of road edge categories involved: Primary, secondary, tertiary, residential, and unclassified (a detailed introduction of this information can be found at \url{https://wiki.openstreetmap.org/wiki/Key:highway}). These types of edge-level geography information are collected from the Open Street Map and encoded into one-hot vectors to represent the edge-level information. We do not use connectivity in the bus network because the available data does not distinguish the direction of routes among the bus stops. Although including this information would complete the heterogeneous map, it is beyond the scope of this research.
    \item \textbf{Labels}. In the land use inference problem, there is no established paradigm for labelling. We adhere to the recommended practices of the most cited foundational studies \cite{2012Inferring} – using the density of POIs to characterise the intensity of specific land use characteristics in a region. To ensure comprehensiveness and objectivity, we selected six POI categories from the OSM database: office, sustenance, transport, retail, leisure and residence. Due to the restriction of OSM's API on the quantity of data collected, it is difficult to obtain complete POI data for the London area. Consequently, we chose a few major tags to represent each land use category, as detailed in Table \ref{tab: category_tags} (the specific introduction of the used tags can be found at \url{https://wiki.openstreetmap.org/}). The spatial distributions of these labels are displayed in Figure \ref{fig: POI-label} in \ref{sec: appendix B}.
    \item \textbf{Catchment area}. 
    We establish a criterion that assigns each node a catchment area—a circular region centred on the node with a predetermined radius. Each node is labelled by the land use indicators within this catchment area. This process considers two types of label tags:
    \begin{itemize}
        \item For point tags (e.g., locations of different services), count the number of target types within the catchment area.
        \item For polygon tags (e.g., parks, buildings), calculate the proportion of the catchment area that these tags cover.
    \end{itemize}
    We have used a 1 km radius for the catchment area in our analysis. As our focus is on demonstrating the use of heterogeneous graph models, a comparison of different catchment radii and their impacts is left to future work.
\end{itemize}

\begin{table}[htbp]
\centering
\small
\caption{POI categories and their specific tags for labelling land use.}
\begin{tabular}{l p{0.5\textwidth} l}
\toprule
\textbf{Category} &\textbf{Tags}& \textbf{Data types} \\
\midrule
Office & Government, Association, Company, Consulting, Financial, 
Coworking, Lawyer, Estate, Insurance, Telecommunication & Points\\
\midrule
Sustenance & Restaurant, Cafe, Bar, Pub, Fast Food & Points\\
\midrule
Transport & Bicycle, Subway, Transport, Taxi, Parking, Ferry, Tram & Points and Polygons\\
\midrule
Retail & Convenience, Supermarket, Mall, Department, Bakery, Butcher, Clothes, Hardware, Furniture, Electronics &  Points\\
\midrule
Leisure & Park, Sports, Playground, Swimming, Pitch, Track, Fitness, Garden, Nature & Points and Polygons\\
\midrule
Residence & Residential, House, Detached, Apartments, Population & Points and Polygons\\
\bottomrule
\end{tabular}
\label{tab: category_tags}
\end{table}

\subsection{Baseline models for comparison}

In our framework, we compare our proposed HGT model against several baseline methods, including:
\begin{itemize}
    \item \textbf{Neural Network (NN):} A basic neural network model, serving as a foundational benchmark against more complex graph-based models.
    \item \textbf{Graph Convolutional Network (GCN):} Utilises convolutional processes directly on the graph structure to capture neighbourhood information \cite{kipf2016semi}.
    \item \textbf{Graph Attention Network (GAT):} Incorporates attention mechanisms into graph neural networks, enabling nuanced weighting of node contributions from neighbours \cite{velivckovic2017graph}.
    \item \textbf{GraphSage:} A scalable graph neural network method that generates embeddings by sampling and aggregating features from a node’s local neighbourhood \cite{hamilton2017inductive}.
    \item \textbf{Relational Graph Convolutional Network (RGCN):} Extends GCNs to manage graph heterogeneity by incorporating various relational types among nodes during the learning process \cite{schlichtkrull2018modeling}.
\end{itemize}

Each of these models represents a unique approach to incorporating graph structure into the learning process, which makes them suitable baselines for assessing the performance of our proposed HGT model.

\subsection{Measurement metrics}

To evaluate the performance of our regression models, we employ three widely recognised metrics: Mean Absolute Error (MAE), Root Mean Square Error (RMSE), and the Coefficient of Determination ($R^2$). These metrics are essential for assessing the accuracy and predictive quality of the models. They are defined as follows,
where $n$ is the number of observations in the test set, $y_i$ represents the actual values, and $\hat{y}_i$ are the predicted values.

\begin{itemize}
    \item \textbf{Mean Absolute Error (MAE):} MAE quantifies the average absolute difference between the predicted values and the actual values, giving an insight into the overall error magnitude without the influence of error direction.
    \[
    \text{MAE} = \frac{1}{n} \sum_{i=1}^n |y_i - \hat{y}_i|
    \]

    \item \textbf{Root Mean Square Error (RMSE):} RMSE is a quadratic scoring rule that calculates the square root of the average of squared differences between the predicted and actual values. It emphasises larger errors, which can be particularly important in certain contexts.
    \[
    \text{RMSE} = \sqrt{\frac{1}{n} \sum_{i=1}^n (y_i - \hat{y}_i)^2}
    \]

    \item \textbf{Coefficient of Determination (R-Squared):} $R^2$ is a statistical measure that indicates the proportion of the variance in the dependent variable that is predictable from the independent variables. It is used to measure how well the model captures the variability of the dataset.
    
    \begin{align*}
      \text{R}^2 = 1 - \frac{\sum_{i=1}^n (y_i - \hat{y}_i)^2}{\sum_{i=1}^n (y_i - \overline{y})^2}  
    \end{align*}
    
    where $\overline{y}$ is the mean of the actual values $y_i$.
\end{itemize}

\subsection{Results}

 These experiments were undertaken on a Linux Ubuntu system, utilising CUDA Version 11.4 with NVIDIA RTX A5000 GPU for deep learning model optimisation. Computations were executed on Python 3.10 within a dedicated virtual environment (Readers can check ``requirement.txt'' on the GitHub website of this project).   For comprehensive model evaluation and testing, we randomly allocated 70\% of the samples for training, with the remaining 15\% for validation and 15\% for testing. 
 To ensure fair comparisons against multiple baseline models, we used the same hyper-parameters to model, train, and test them, as detailed in Table \ref{tab:experiment_settings}.

\begin{table}[ht]
\centering
\caption{Summary of Hyper-Parameter Settings.}
\label{tab:experiment_settings}
\begin{tabular}{@{}lc@{}}
\toprule
\textbf{Parameter}                      & \textbf{Value}           \\ \midrule
Hidden Layer Numbers                    & 2                        \\
Hidden Layer Dimension                  & 128                      \\
Activation Function
 & ReLU                     \\
Attention-Head Numbers                  & 2                        \\
Optimiser                               & AdamW \cite{loshchilov2017decoupled}                    \\
Epoch Number                            & 200                      \\
Batch-Size                              & 64                       \\ 
Learning Rate                           & 0.002                    \\
Sampling for Homo-Graph      & Neighbor Loader
\cite{hamilton2017inductive}     \\
Sampling for Hetero-Graph    & HGT Loader \cite{hu2020heterogeneous}          \\
Graph Hop Number                              & 1                        \\
\bottomrule
\end{tabular}
\end{table}
 
\subsubsection{Overall accuracy}
\label{ssec:overall_accuracy}
This section offers a comprehensive evaluation of various deep learning methods using the London mobility datasets across multiple urban domains. Models evaluated include the proposed HGT framework compared with baseline methods mentioned in Section 5.2. The improvement percentage (Impr.) over the baseline neural network model underscores the effectiveness of these models in inferring land use.

\begin{table}[htbp]
\centering
\small 
\caption{Performance metrics and improvements across different land use targets.}
\setlength{\tabcolsep}{4pt} 
\renewcommand{\arraystretch}{1.1} 
\begin{tabular}{llcccccc}
\toprule
\textbf{Indicators} & \textbf{Metrics} & \textbf{NN} & \textbf{GCN} & \textbf{GraphSage} & \textbf{GAT} & \textbf{RGCN} & \textbf{HGT} \\
\midrule
\multirow{4}{*}{Office} & MAE & 0.086 & 0.043 & 0.043 & 0.039 & 0.035 & \textbf{0.031} \\
 & Impr. & - & 50.00\% & 50.00\% & 54.65\% & 59.30\% & \textbf{63.95\%} \\
 \cmidrule{2-8}
 & RMSE & 0.124 & 0.060 & 0.058 & 0.054 & 0.051 & \textbf{0.047} \\
 & Impr. & - & 51.61\% & 53.23\% & 56.45\% & 58.87\% & \textbf{62.10\%} \\
\cmidrule{1-8}
\multirow{4}{*}{Sustenance} & MAE & 0.079 & 0.033 & 0.032 & 0.034 & 0.031 & \textbf{0.025} \\
 & Impr. & - & 58.23\% & 59.49\% & 56.96\% & 60.76\% & \textbf{68.35\%} \\
 \cmidrule{2-8}
 & RMSE & 0.116 & 0.046 & 0.045 & 0.049 & 0.044 & \textbf{0.039} \\
 & Impr. & - & 60.34\% & 61.21\% & 57.76\% & 62.07\% & \textbf{66.38\%} \\
\cmidrule{1-8}
\multirow{4}{*}{Transport} & MAE & 0.097 & 0.046 & 0.043 & 0.039 & 0.039 & \textbf{0.033} \\
 & Impr. & - & 52.58\% & 55.67\% & 59.79\% & 59.79\% & \textbf{65.98\%} \\
 \cmidrule{2-8}
 & RMSE & 0.139 & 0.063 & 0.059 & 0.054 & 0.054 & \textbf{0.050} \\
 & Impr. & - & 54.68\% & 57.55\% & 61.15\% & 61.15\% & \textbf{64.03\%} \\
\cmidrule{1-8}
\multirow{4}{*}{Retail} & MAE & 0.103 & 0.057 & 0.053 & 0.050 & 0.046 & \textbf{0.039} \\
 & Impr. & - & 44.66\% & 48.54\% & 51.46\% & 55.34\% & \textbf{62.14\%} \\
 \cmidrule{2-8}
 & RMSE & 0.134 & 0.077 & 0.070 & 0.068 & 0.065 & \textbf{0.058} \\
 & Impr. & - & 42.54\% & 47.76\% & 49.25\% & 51.49\% & \textbf{56.72\%} \\
\cmidrule{1-8}
\multirow{4}{*}{Leisure} & MAE & 0.106 & 0.064 & 0.060 & 0.058 & 0.051 & \textbf{0.044} \\
 & Impr. & - & 39.62\% & 43.40\% & 45.28\% & 51.89\% & \textbf{58.49\%} \\
 \cmidrule{2-8}
 & RMSE & 0.141 & 0.086 & 0.083 & 0.080 & 0.069 & \textbf{0.067} \\
 & Impr. & - & 39.01\% & 41.13\% & 43.26\% & 51.06\% & \textbf{52.48\%} \\
\cmidrule{1-8}
\multirow{4}{*}{Residence} & MAE & 0.123 & 0.062 & 0.058 & 0.054 & 0.051 & \textbf{0.042} \\
 & Impr. & - & 49.59\% & 52.85\% & 56.10\% & 58.54\% & \textbf{65.85\%} \\
 \cmidrule{2-8}
 & RMSE & 0.158 & 0.083 & 0.078 & 0.074 & 0.070 & \textbf{0.062} \\
 & Impr. & - & 47.47\% & 50.63\% & 53.16\% & 55.70\% & \textbf{60.76\%} \\
\bottomrule
\end{tabular}
\label{tab:metrics_improvement}
\end{table}

The results in Table \ref{tab:metrics_improvement} 
indicate a consistent improvement in performance when transitioning from general deep learning models to complex graph-based models. The HGT model consistently delivers the highest improvement percentages across all categories, showing its capability to handle heterogeneous graph data. Specifically, the `Sustenance' indicator has substantial benefits from graph-based models, with the HGT achieving MAE improvements of 68.35\%. The RGCN and HGT models consistently outperform others, probably due to their ability to capture heterogeneous relationships within the mobility and topology structure. The smallest improvement is noted in the `leisure' land use indicator, suggesting that there might be unobserved information which could boost predictive accuracy.

\subsubsection{Spatial residual distribution}

\begin{figure}[ht]
    \centering
    \includegraphics[width=1\linewidth]{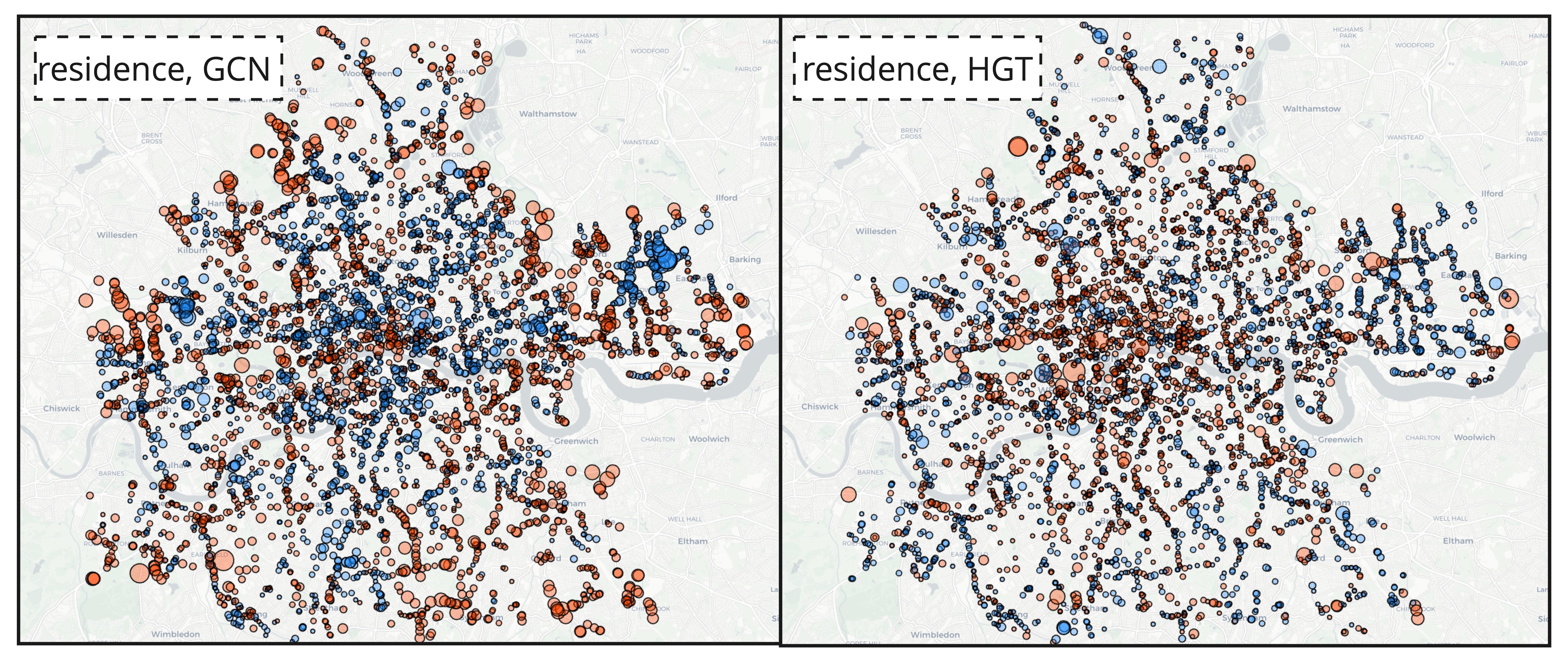}
    \caption{Spatial distribution of residual of residence: GCN (left) and HGT (right)}
    \label{fig: residual map}
\end{figure}

To better understand the function of heterogeneous graph models at the individual level, we compare the geographical distribution of residuals in the geometric map between GCN and HGT. Figure \ref{fig: residual map} illustrates an example of one of the land use indicators –`Residence', where the size of the circular markers indicates the absolute value of the residuals. The colour of each marker indicates whether the residual is positive or negative: red for over-predicted values and blue for underestimated values. Similar comparisons of other land use indicators can be found in Figure \ref{fig: all residual map} in \ref{sec: appendix C}.

A comparison in Figure \ref{fig: residual map} reveals that the circular markers in HGT are smaller overall than those in GCN. Additionally, a continuous distribution of red and blue colours in the GCN map suggests that the model consistently overestimates or underestimates certain areas or roads. This suggests unobserved heterogeneity is hindering the model's performance. HGT significantly improves this issue, displaying a more uniform colour distribution in its corresponding maps. According to Figure \ref{fig: all residual map}, the same observations 
applied
in other land use indicators like `leisure' and `retail'.

\subsubsection{Ablation analysis on connectivity in the graph}

\begin{figure}[ht]
    \centering
    \includegraphics[width=0.8
    \linewidth]{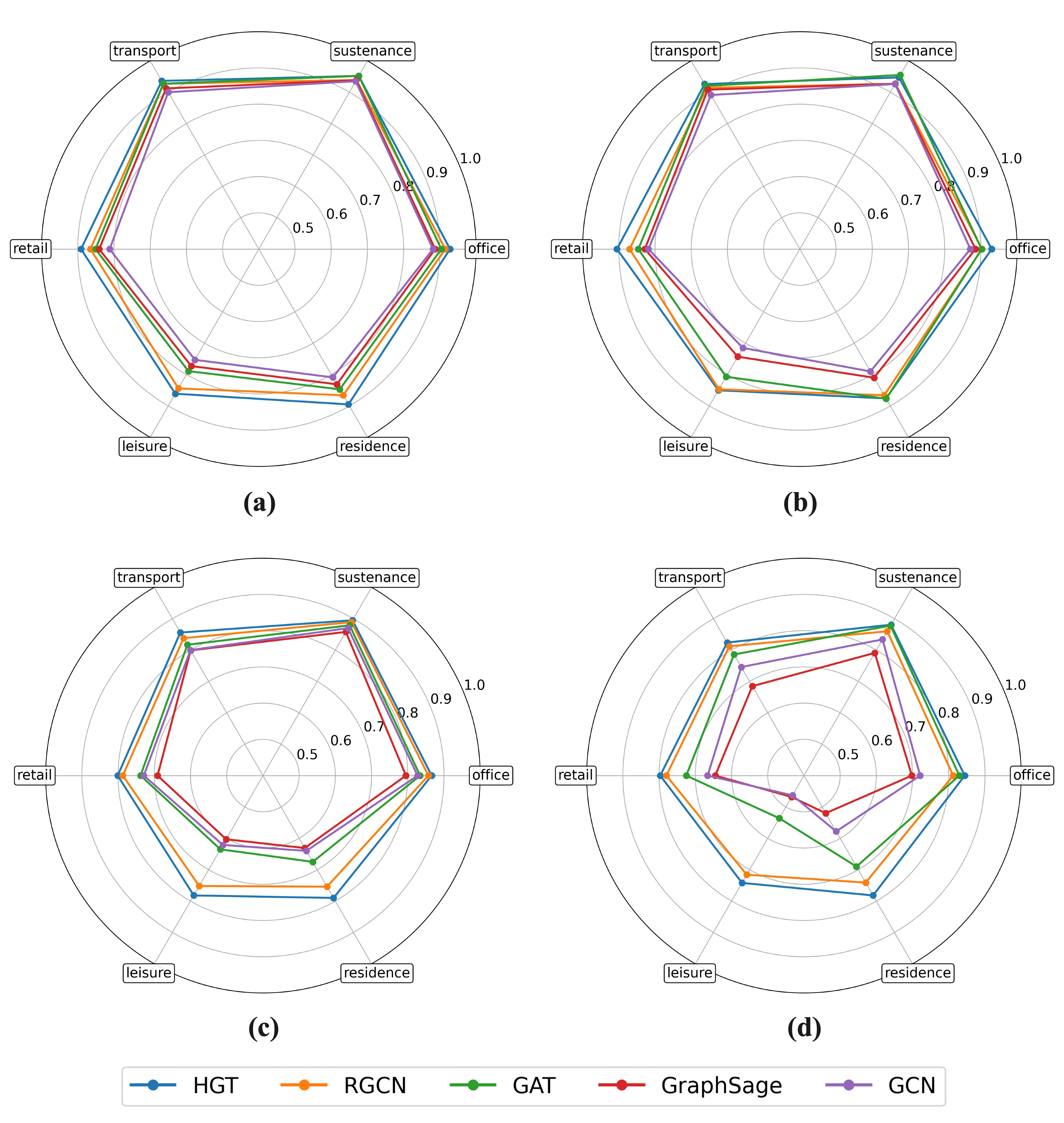}
    \caption{$R^2$ of six land use indicators in four scenarios: (a) bus, bike, and tube; (b) bus and bike; (c) bus and tube; (d) bus}
    \label{fig: R2-radar}
\end{figure}

Given the significant improvement of GNN models illustrated in Table \ref{tab:metrics_improvement}, this section delves into the role of connectivity and neighbouring nodes. Specifically, we constructed four scenarios by removing some of the nodes and connections: a) all nodes and edges; b) only bus and bike nodes with their edges; c) only bus and tube nodes with their edges; d) only bus nodes with their edges. Owing to the limited number of bike and tube nodes, these modes were not assessed as separate scenarios. We use R-square as the measurement for comparison as shown in Figure \ref{fig: R2-radar}. 

According to Figure \ref{fig: R2-radar}, we can 
draw 
two conclusions. First, increased connectivity enhances performance across all metrics in all graph models. Specifically, moving from chart (a) to chart (d) reveals a noticeable decrease in performance metrics across domains like transport, retail, and residential areas. Second, the heterogeneous information acts as complementary for prediction, especially for areas under sparse connectivity. As noted previously, inferring land use for leisure is challenging because vehicular and cycling mobility data poorly represent non-commuting activities like leisurely travelling. However, we observed that heterogeneous graph models (HGT and RGCN) significantly improve the accuracy performance in the `leisure' land use. 

\subsubsection{Feature attribution analysis}

\begin{figure}[ht]
    \centering
    \includegraphics[width=1\linewidth]{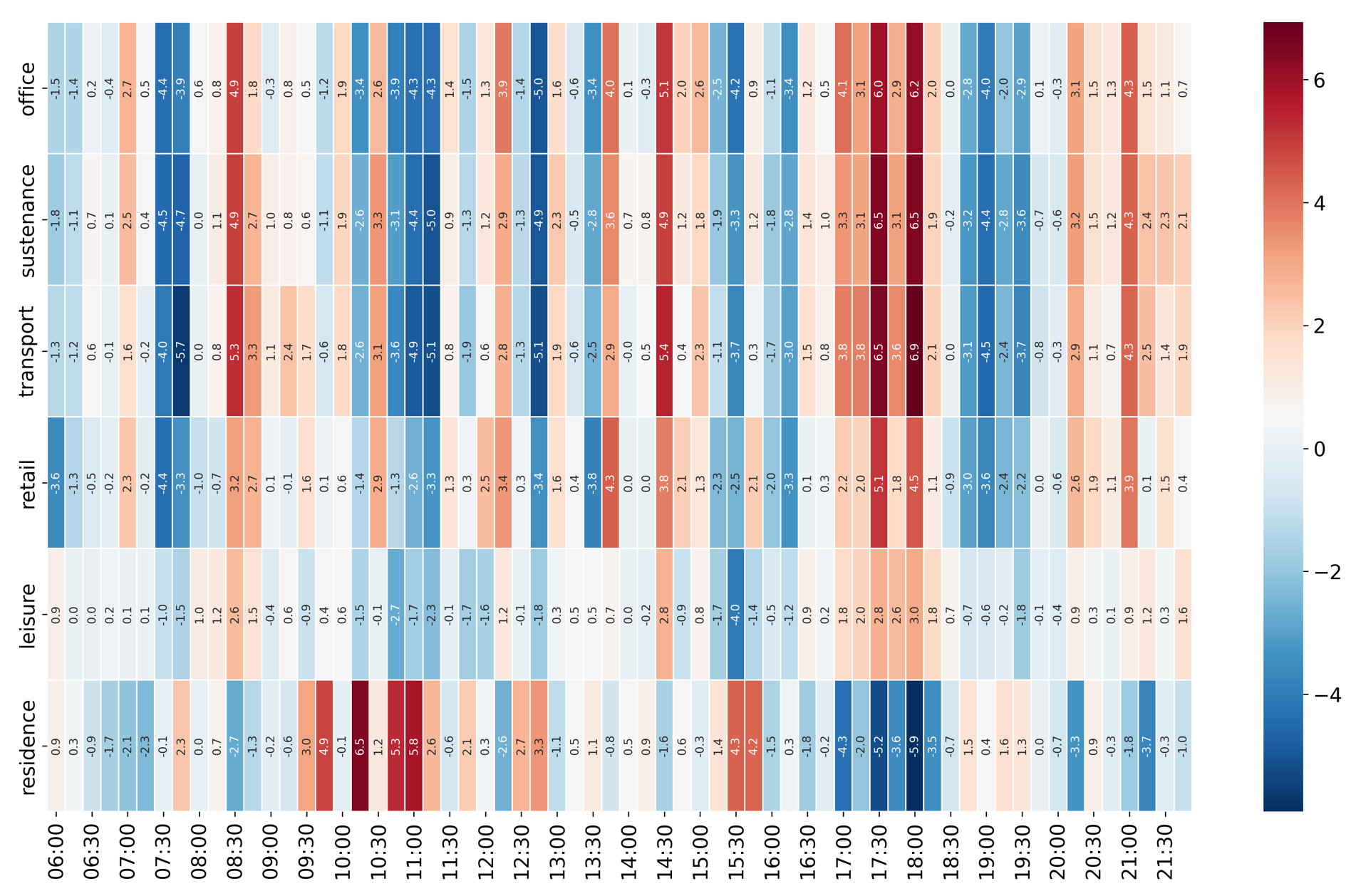}
    \caption{Heatmap of the distribution of Integrated-gradients importance of HGT model.}
    \label{fig: heatmap}
\end{figure}

This section examines feature attributions – the impact of the daily mobility distribution on the outcome using the Approximate Integrated Gradients index. The results are displayed in a heatmap format, where red indicates a positive correlation and blue a negative correlation between input features of different 15-time intervals and land use indicators, as shown in Figure \ref{fig: heatmap}. 

The feature attribution results in Figure \ref{fig: heatmap} closely align with the general understanding of human activities in London, illustrated in three categories with different characteristics. The first category, comprising \textbf{transport, office, and sustenance}, is primarily associated with work-related activities. This category shows a high sensitivity to commuter inflow during peak hours. Specifically, inflow is lower in the morning (10:00-11:00 AM) and significantly higher in the evening (5:00-6:00 PM). This indicates that areas with these land uses attract more commuters returning from work in the evening than departing in the morning. Besides, a subtler trend shows that Category 1 (work-related) stations experience reduced feature importance in the early evening (6:45-7:45 PM) and increased importance later at night (8:45-9:45 PM). This likely reflects a common urban lifestyle pattern, where individuals engage in leisure and entertainment activities after returning home from work.

In contrast, the second category, consisting of \textbf{residential} land use areas, exhibits a reverse trend with higher morning inflow and lower evening inflow. This reflects typical residential patterns where people leave for work in the morning and return in the evening.

The third category, including \textbf{leisure and retail}, is less affected by inflows at peak hours. Notably, these areas show minimal variation in inflow during peak hours but have varied activity during other times, such as evenings (presumably after work), which aligns with shopping and leisure activities.

It is important to note that various feature attribution methods might produce different outcomes. Our comparison of two methods, Integrated Gradients (Figure~\ref{fig: heatmap}) and InputX-Gradient (Figure~\ref{fig: heat map by inputX} in appendix), illustrates this point. Integrated Gradients are generally considered more reliable because they account for changes over a broader range, offering a more comprehensive understanding of feature importance. Conversely, InputX-Gradient, focusing on a single point gradient, failed to align with the general context. As shown in this case, the importance of heatmap from the Integrated Gradient method aligns more closely with the characteristics of human activity in London, reflecting our understanding of the comparative performance of the methods. 

\subsubsection{Counterfactual explanation analysis}
\label{ssec:counterfactual_analysis}

In the previous section, we focused on the explainability of our model by analysing the feature attribution of node features. Although this analysis offered valuable insights into individual node attributes,  it did not account for the importance of other critical factors such as node types, edge types, and the overall graph structure. To overcome this limitation, this section introduces counterfactual analysis for a more comprehensive understanding of our model.
We set `mixed' as an expected scenario. Mixed land use refers to an area with balanced facilities and diverse activities, crucial for enhancing urban vitality, reducing commute distances, and promoting sustainable development \cite{lehmann2016sustainable}. Mathematically, given the predicted land use intensity $y_{i,j} 
$ from the main models, we applied the Shannon diversity index to quantify the diversity of land uses within each area. The Shannon diversity (SD) is calculated as follows:

\begin{equation}
\text{SD}_i = -\sum_{j\in \mathcal{T}} \frac{y_{i,j}}{\sum_{j \in \mathcal{T}}y_{i,j}} \ln(\frac{y_{i,j}}{\sum_{j\in \mathcal{T}}y_{i,j}})
\end{equation}
where \( y_{i,j}/\sum_{j\in \mathcal{T}} y_{i,j} \) represents the proportion of land use of type \( i \) in the area. This formula provides a diversity score for each area to reflect the variety of land uses present. After calculating the diversity scores for all areas, we ranked them in descending order. The areas with the highest diversity scores, representing the top 10\% of all nodes, were classified as `mixed' land use types. This approach ensures that we focus on regions that already demonstrate a high level of land use diversity.

Based on the mixed land use based Counterfactual Explanations (CEs), this counterfactual analysis is conducted in two parts. Initially, we perform a local-level analysis to demonstrate the practical application of CEs. In this analysis, we select a specific input node and target a `mixed' land use prediction as the desired outcome. We identify a 
CE for this node and compare it with the original input to underscore the necessary changes for achieving the target prediction. Secondly, we conduct an aggregated-level analysis to demonstrate how dissimilarity measures for 
CEs can illuminate the importance of factors such as node features, node types, edge types, and graph structure. 

\begin{figure}
    \centering
    \includegraphics[width=1\linewidth]{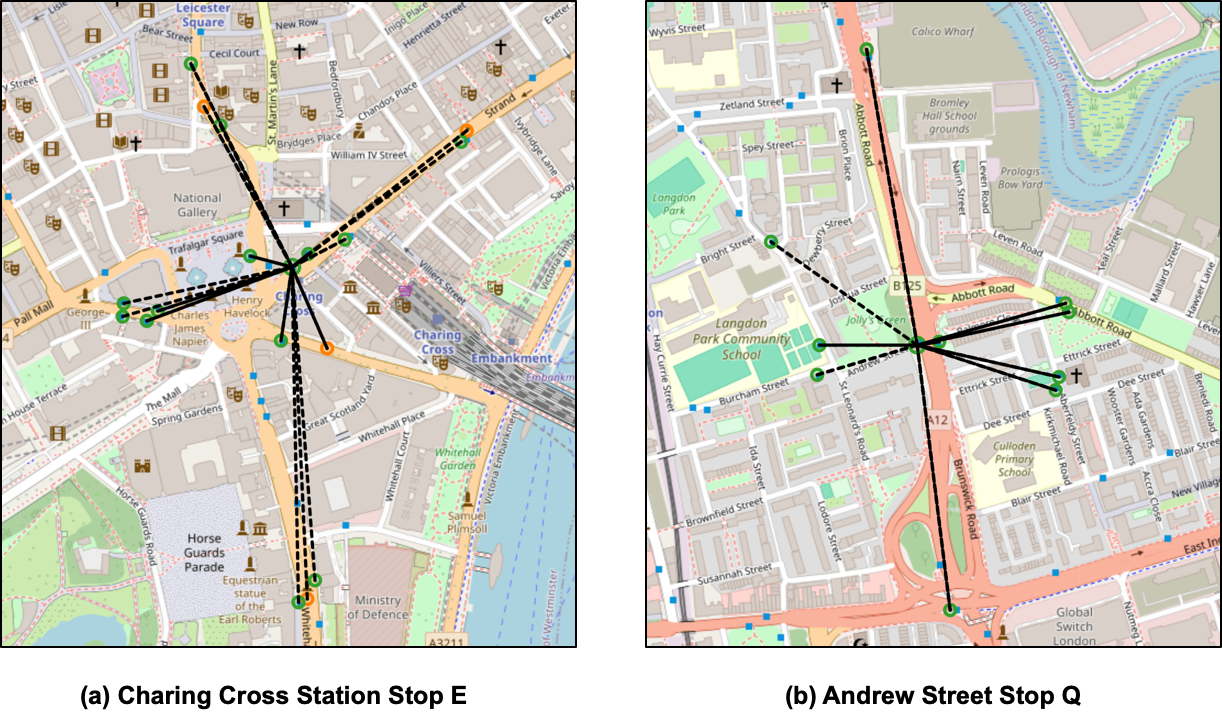}
    \caption{Two local-level cases: (a) Input 1: a bus stop in the central London area with a high level of `transport' prediction; (b) Input 2: a bus stop in the eastern area with a high level of `leisure' prediction}
    \label{fig: local_map}
\end{figure}
\begin{figure*}[ht]
	\centering
\begin{subfigure}{0.48\columnwidth}
    \includegraphics[width=\linewidth]{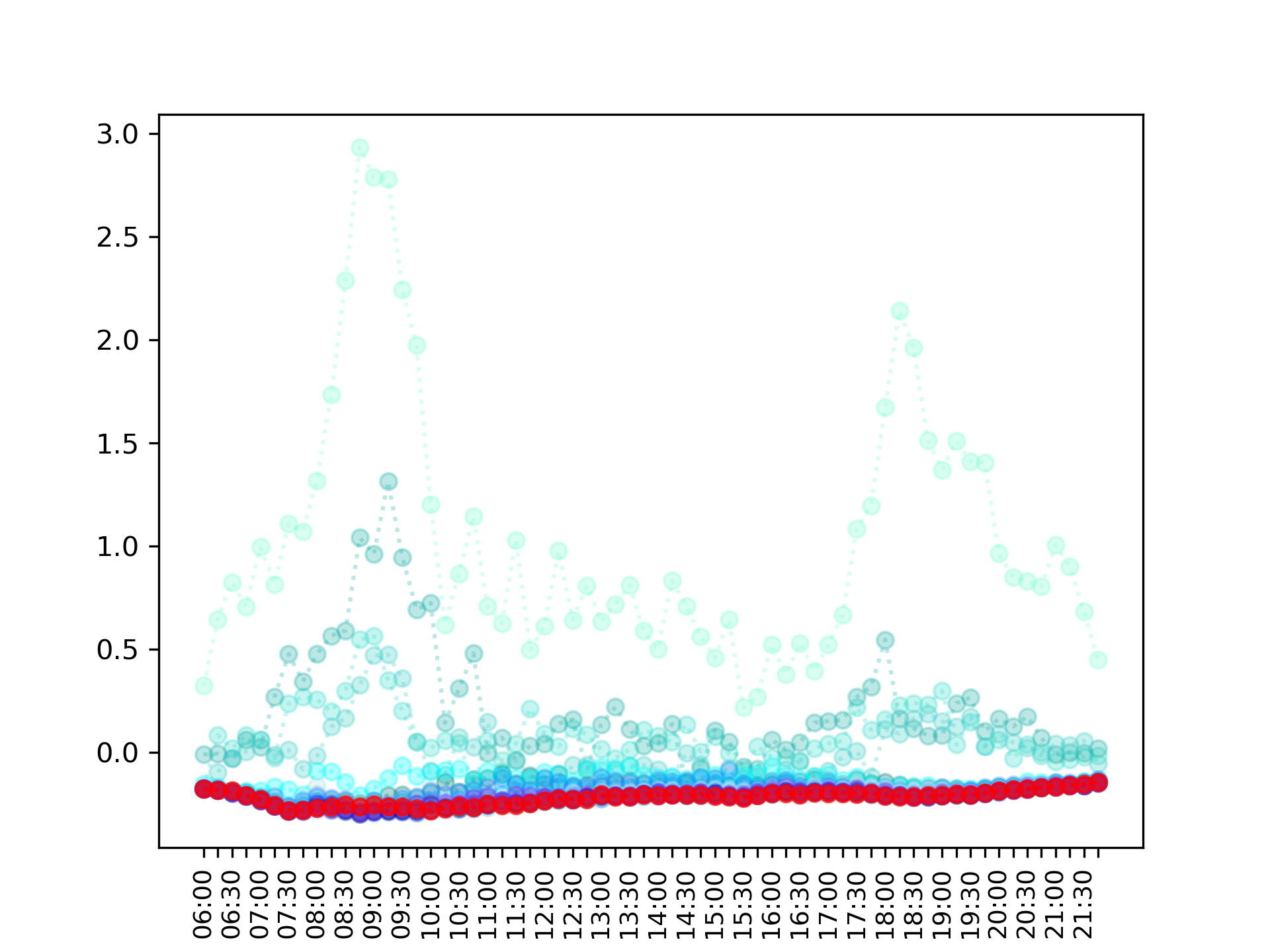}
    \caption{Input 1: Charing Cross stop E, and its CE}
    \label{fig:ce_node_1}
  \end{subfigure}
  \begin{subfigure}{0.48\columnwidth}
    \includegraphics[width=\linewidth]{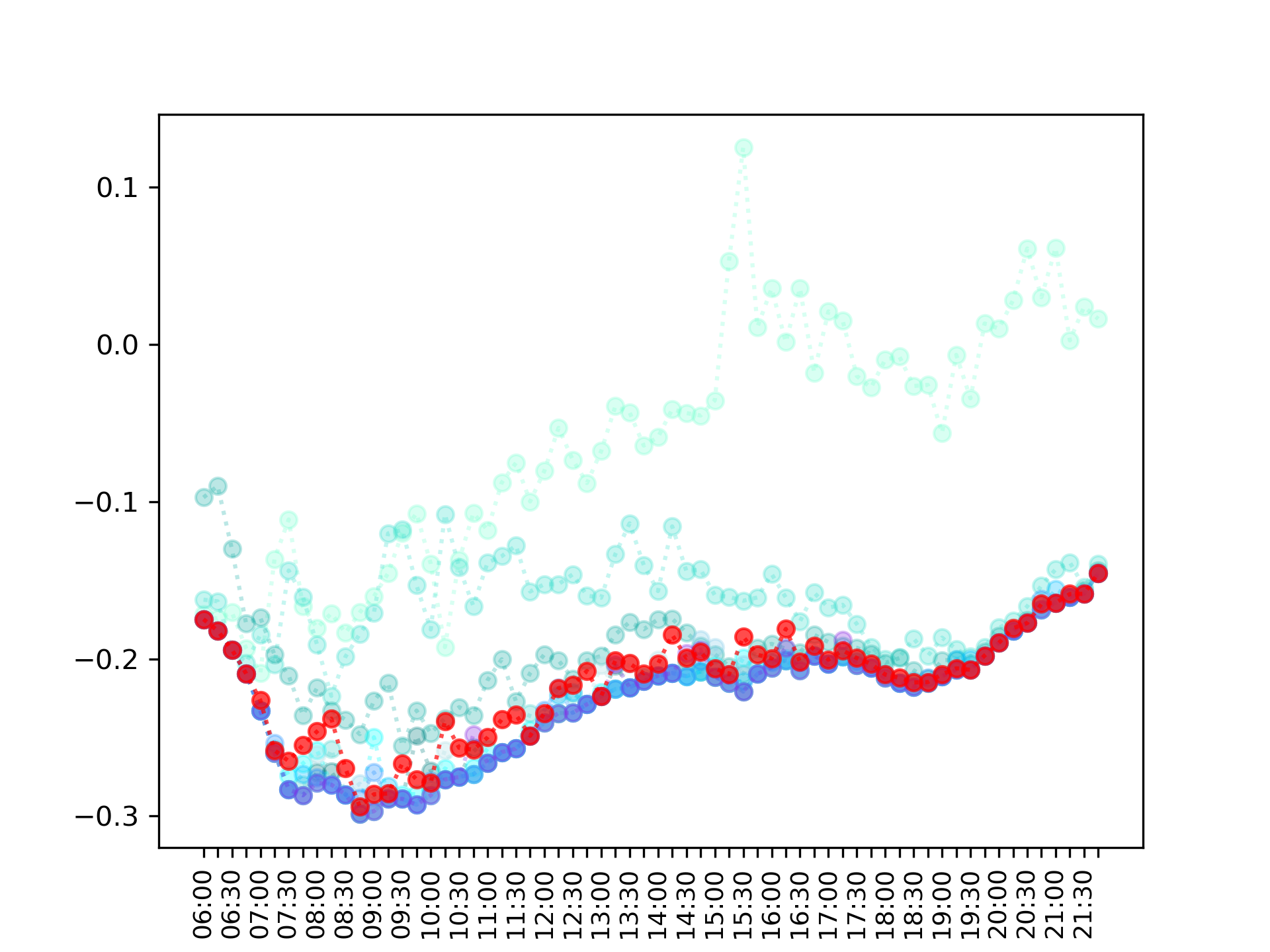}
    \caption{Input 2: Andrew Street Stop Q, and it CE}
    \label{fig:ce_node_2}
  \end{subfigure}
 \caption{Node feature differences. Each line (in different colours) is a vector of node features of a node, the line highlighted in red is for the input node, while the other lines in blue-like colours are for the nodes in the CE subgraph. }
\label{fig:counterfactual_examples_plots}
\end{figure*}

\paragraph{Local-level analysis} 
We present CEs for two example input nodes which do not belong to the `mixed' land use area; these are  the inputs 1 and 2 
which are located in Central and East London respectively (Figure~\ref{fig: local_map}). We compute CEs for the two nodes which reveal the minimum changes to make from the input (subgraph) in order to achieve the `mixed' land use. The node feature differences between the input nodes and the CE subgraphs are illustrated in Figure~\ref{fig:counterfactual_examples_plots}, and the node type, edge type, and graph structure differences are listed in Table~\ref{tab:counterfactual_examples}. For input 1, as can be observed from the magnitudes in Figure~\ref{fig:ce_node_1}, the required changes are mostly in the node features, while the differences are small for the items in Table~\ref{tab:counterfactual_examples}. Indeed, this intuition is demonstrated by the dissimilarity measures, which for the four data modalities are respectively 0.271, 0.056, 0.043, and 0.043. The high node feature dissimilarity emphasises the importance of changing node features for input 1. For input 2, however, the dissimilarities are 0.047, 0.0, 0.133, and 0.069, meaning that the change of edge (road) type is the primary factor in achieving the `mixed' land use. The magnitudes of node feature changes are relatively small (Figure~\ref{fig:ce_node_2}), while the edge type changes are more significant.


\begin{table}[ht]
    \small
    \centering
    \setlength{\tabcolsep}{4pt} 
    \caption{Node type, edge type, graph structure differences between example inputs and their CE, both in the form of 1-hop complete subgraph.}
    \begin{tabular}{cccc}
        \toprule
        \textbf{Subgraph} & \textbf{Node type} & \textbf{Avg. Edge Type} & \textbf{Structure}\\
        \midrule
        Input 1 & 0 tube, 14 bus, 4 bike & 0.52 & 141  \\
        CE 1 & 0 tube, 13 bus, 4 bike & 0.70 & 135 \\
        \midrule
        Input 2 & 0 tube, 11 bus, 0 bike & 1.68 & 72  \\
        CE 2 & 0 tube, 11 bus, 0 bike & 1.15 & 67 \\
        \bottomrule
    \end{tabular}
    \label{tab:counterfactual_examples}
\end{table}

\paragraph{Aggregated-level analysis} Given a closed `mixed' subgraph for each sample's subgraph, we specify four types of differences: node feature, node type, edge type and graph structure. These differences can be regarded as relevant importance determining the minimal effort required to transition to a mixed land use type. All the values have been scaled into the 0-1 range, as shown in Table \ref{tab:counterfactual_scores}.

\begin{table}[ht]
    \small
    \centering
    \setlength{\tabcolsep}{4pt} 
    \caption{Counterfactual scores of different elements in a graph. }
    \begin{tabular}{lcccc}
        \toprule
        \textbf{Indicators} & \textbf{Node feature} & \textbf{Node type} & \textbf{Edge type} & \textbf{Structure} \\
        \midrule
        office & 0.637 $\pm$ 0.132 & 0.422 $\pm$ 0.139 & 0.030 $\pm$ 0.030 & 0.021 $\pm$ 0.023 \\
        sustenance & 0.449 $\pm$ 0.212 & 0.457 $\pm$ 0.089 & 0.049 $\pm$ 0.045 & 0.039 $\pm$ 0.033 \\
        transport & 0.287 $\pm$ 0.234 & 0.370 $\pm$ 0.121 & 0.038 $\pm$ 0.034 & 0.048 $\pm$ 0.052 \\
        retail & 0.262 $\pm$ 0.302 & 0.399 $\pm$ 0.130 & 0.033 $\pm$ 0.021 & 0.042 $\pm$ 0.038 \\
        leisure & 0.284 $\pm$ 0.260 & 0.269 $\pm$ 0.143 & 0.058 $\pm$ 0.044 & 0.048 $\pm$ 0.046 \\
        residence & 0.216 $\pm$ 0.202 & 0.286 $\pm$ 0.141 & 0.045 $\pm$ 0.039 & 0.048 $\pm$ 0.052 \\
        \bottomrule
    \end{tabular}
    \label{tab:counterfactual_scores}
\end{table}

According to Table \ref{tab:counterfactual_scores}, In terms of the average values, node feature and node type are the primary source of discrepancies, while edge type and graph structure have lower impact. Specifically, in leisure and residence tasks, the impact of node features is significantly reduced, which is aligned with the relatively low accuracy of these two tasks, as shown in Section~\ref{ssec:overall_accuracy}. Conversely, these tasks slightly increase the influence of edge types and graph structure. 
Additionally, the standard deviation indicates that node features contribute minimally to the difference for some sample regions, especially for indicators like `retail', `leisure' and `residence'. These patterns indicate that although node features are generally crucial, graph structural aspects and node relationships become more significant in leisure and residence contexts.
\section{Conclusions}
\label{sec:conc}

This paper builds an explainable graph-based framework for inferring land use from multi-modal mobility data in complex urban scenarios. In this framework, a Heterogeneous Graph Transformer model is introduced to fuse multi-modal mobility information for land use inference tasks. Next, two types of explainable AI methods enhance the models' transparency and extrapolability. 

The overall accuracy indicates that the use of GNNs significantly improves accuracy. Furthermore, ablation analysis shows that HGNs effectively enhance network structure information and outperform homogeneous graph models, especially in challenging tasks like identifying `leisure'. 

Feature attribution analysis assesses the importance of input factors and verifies the model's explainability by checking if the importance distribution aligns with expert experience. As shown in the feature attribution analysis, the symmetrical nature of the `residence’ and `work’ categories aligns well with the commuter's `work'; and `recreation' activities in London, 
giving some confidence in the suitability of proposed model for the land use setting. Meanwhile, counterfactual explanations highlight the efforts needed to alter any node's prediction from a micro perspective, allowing experts to assess the GNNs' explainability based on differences between subgraphs. The analysis of the counterfactual explanations in Section~\ref{ssec:counterfactual_analysis} reveals that in areas with different functions and historical backgrounds (e.g., Central London and East London), the different types of information in the heterogeneous graph data, which represent various elements in the sub-map, have significantly different forms of contribution. This is consistent with the planner's understanding of these regions, thus validating the usefulness of fusing data from different sources. We posit that these insights will help to build user trust in GNNs by providing a clear, evidence-based rationale for each prediction, which is crucial for users in urban planning and management to trust the proposed methods.


 There are several limitations and opportunities for future research. The current model uses simplistic meta-relations that consider only 1-hop relationships. Future work could focus on more complex relationships (e.g., multi-steps of transfers) among urban mobility facilities. Due to the limited data collected, the nodes' features are only represented by outflow rather than inflow and outflow, potentially limiting the comprehensiveness of our findings. Future research will benefit from the growing availability of open graph-structured urban data, enabling comparative studies to enhance our understanding of urban dynamics in different countries and societies. Cross-validation could potentially enhance this model's performance; however, due to limited hardware computational capacity, only standard training methods were utilised in this paper. Further, incorporating multiple explainable AI techniques will increase the model's transparency and strengthen support for urban planning decisions, offering a more trustworthy analysis of urban land use dynamics.

\section{Declaration of generative AI and AI-assisted technologies in the writing process}

During the preparation of this work, the author(s) used Chatgpt in order to improve readability. After using this tool/service, the author(s) reviewed and edited the content as needed and take(s) full responsibility for the content of the publication.

\appendix

\section{Labels processing}
\label{sec: appendix B}

\begin{figure}
    \centering
    \includegraphics[width=1\linewidth]{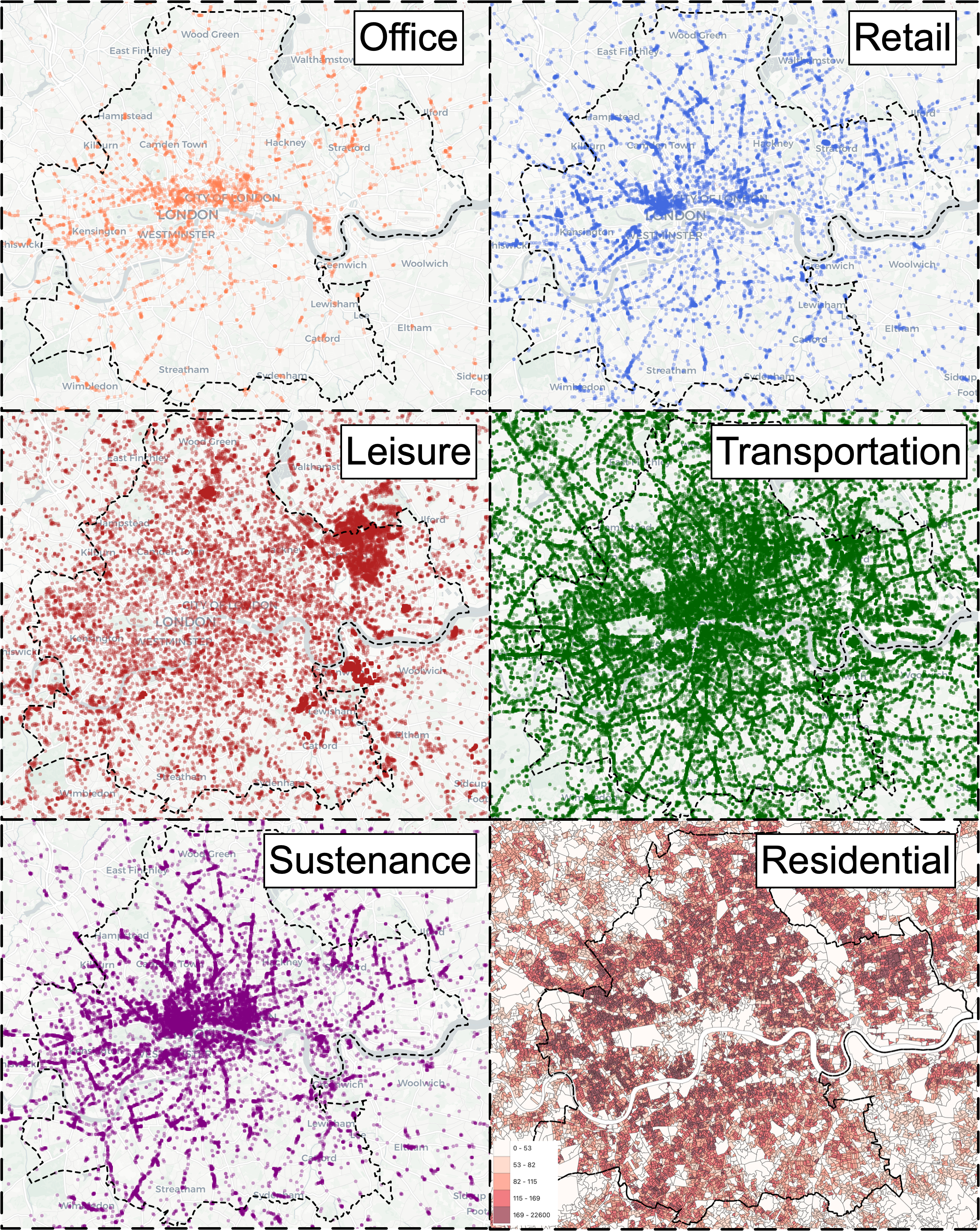}
    \caption{POI and population data for building the labels}
    \label{fig: POI-label}
\end{figure}

\section{Supplementary results}
\label{sec: appendix C}

\begin{figure}
    \centering
    \includegraphics[width=1\linewidth]{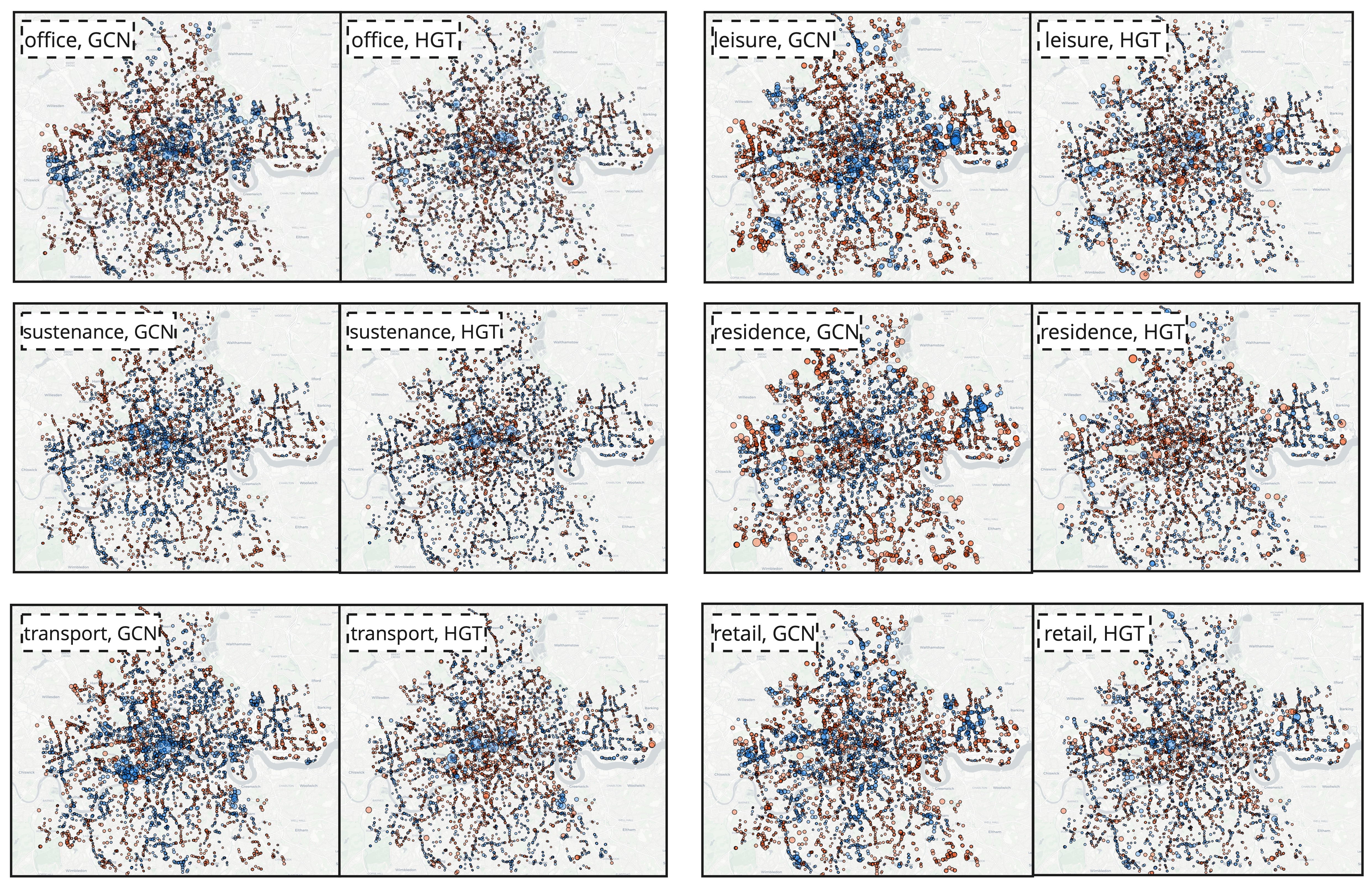}
    \caption{Special distribution of residual among six land use indicators}
    \label{fig: all residual map}
\end{figure}

\begin{figure}
    \centering
    \includegraphics[width=1\linewidth]{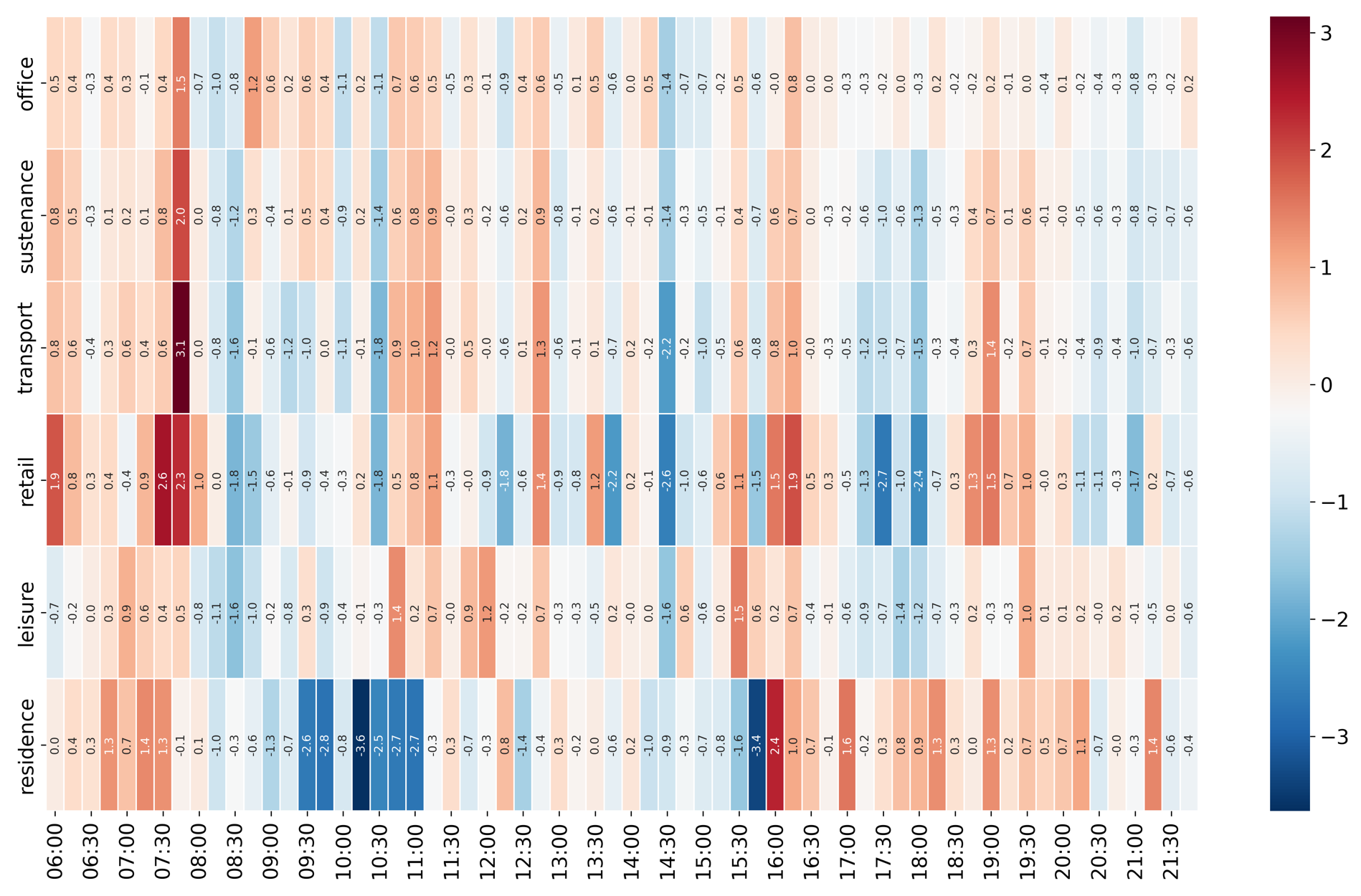}
    \caption{Heatmap of the distribution of InputX-gradient}
    \label{fig: heat map by inputX}
\end{figure}


 \bibliographystyle{elsarticle-num} 
 \bibliography{manuscript}





\end{document}